\documentclass{article}

\usepackage{microtype}
\usepackage{graphicx}
\usepackage{subfigure}
\usepackage{booktabs} 

\usepackage{hyperref}

\usepackage[accepted]{icml2025}

\usepackage{amsmath}
\usepackage{amssymb}
\usepackage{mathtools}
\usepackage{amsthm}

\usepackage[capitalize,noabbrev]{cleveref}

\theoremstyle{plain}
\newtheorem{theorem}{Theorem}[section]

\newtheorem{corollary}[theorem]{Corollary}
\theoremstyle{definition}

\theoremstyle{remark}

\usepackage[textsize=tiny]{todonotes}

\usepackage{multirow}
\usepackage{tablefootnote}
\usepackage{subfigure}
\usepackage{wrapfig}

\DeclareMathOperator*{\argmax}{arg\,max}
\DeclareMathOperator*{\argmin}{arg\,min}
\newcommand{\KL}{D_{\mathrm{KL}}}

\icmltitlerunning{LLM Bandit}

\begin{document}

\twocolumn[
\icmltitle{LLM Bandit: Cost-Efficient LLM Generation via \\Preference-Conditioned Dynamic Routing}



\icmlsetsymbol{equal}{*}

\begin{icmlauthorlist}
\icmlauthor{Yang Li}{}
\end{icmlauthorlist}


\icmlcorrespondingauthor{Yang Li}{yli.ml.research@gmail.com}

\icmlkeywords{Machine Learning, ICML}

\vskip 0.3in
]



\printAffiliationsAndNotice{}  


\begin{abstract}
The rapid advancement in large language models (LLMs) has brought forth a diverse range of models with varying capabilities that excel in different tasks and domains. However, selecting the optimal LLM for user queries often involves a challenging trade-off between accuracy and cost, a problem exacerbated by the diverse demands of individual queries. In this work, we present a novel framework that formulates the LLM selection process as a multi-armed bandit problem, enabling dynamic and intelligent routing of queries to the most appropriate model. Our approach incorporates a preference-conditioned dynamic routing mechanism, allowing users to specify their preferences at inference time, thereby offering a customizable balance between performance and cost. Additionally, our selection policy is designed to generalize to unseen LLMs, ensuring adaptability to new models as they emerge. Experimental results demonstrate that our method achieves significant improvements in both accuracy and cost-effectiveness across various LLM platforms, showcasing the potential of our framework to adaptively optimize LLM selection in real-world scenarios.
\end{abstract}

\section{Introduction}
The rapid advancement in large language models (LLMs) has revolutionized natural language processing, bringing forth a diverse ecosystem of models with varying capabilities and cost profiles. While larger models like GPT-4 demonstrate superior reasoning and generation abilities, they come with substantial costs — often \$0.03-0.10 per query — making them impractical for large-scale deployments \citep{achiam2023gpt}. In contrast, open-source models like Mixtral-8x7B offer competitive performance at roughly 1/10th the cost \citep{jiang2024mixtral}. Additionally, domain-specialized models (e.g., CodeLlama for programming \citep{roziere2023code}, Med-PaLM for healthcare \citep{singhal2023large}) often outperform general-purpose models in their specific domains while maintaining lower operational costs. This diversity creates a complex decision space for organizations deploying LLM applications, where optimal model selection must balance performance, cost, and domain-specific requirements.

Existing approaches to address the performance-cost dilemma typically fall into three categories. First, ensemble methods \citep{jiang2023llm,wang2023fusing} combine responses from multiple LLMs to enhance reliability. While effective, these methods require invoking multiple models per query, multiplying costs and latency. Second, cascading approaches like FrugalGPT \citep{chen2023frugalgpt} and AutoMix \citep{madaan2023automix} implement a sequential strategy, starting with cheaper models and escalating to more expensive ones only when necessary. However, this can lead to increased latency for complex queries that require multiple model invocations. Third, direct routing approaches \citep{ding2024hybrid,ong2024routellm,nguyen2024metallm} attempt to select the most appropriate model without multiple invocations. While more cost-efficient, current routing systems struggle with generalization and adaptation to new models.

Designing an effective routing system for real-world deployment presents several fundamental challenges. First, LLMs encounter diverse queries ranging from simple factual questions to complex reasoning tasks. For instance, while a smaller model might handle basic customer service queries adequately, it may struggle with technical documentation generation. The routing mechanism must effectively assess query complexity and model capabilities. Second, the LLM landscape evolves rapidly, with new models released monthly. A routing system must adapt to new models without extensive retraining. Third, different applications have varying requirements — a customer service chatbot might prioritize response speed and cost efficiency, while a legal document analysis system might require higher accuracy regardless of cost. The routing mechanism must dynamically adjust to these varying preferences. Finally, the routing decision itself must be lightweight, adding minimal overhead to the overall query processing pipeline.

To address these challenges, we propose a preference-conditioned dynamic routing mechanism that frames the LLM selection process as a multi-armed bandit problem. Our approach introduces several key innovations:  First, we develop model identity vectors that capture each model's capabilities across different tasks and domains, enabling efficient comparison and selection. Second, users can specify their desired trade-off between performance and cost at inference time through a simple preference parameter, allowing dynamic adaptation to different use cases. Third, new models can be quickly incorporated by evaluating their performance on a carefully selected subset of benchmark prompts, typically requiring only 20-50 evaluations.

Our contributions are significant in both theoretical and practical aspects. We formulate the routing problem as a multi-objective optimization task, balancing the trade-off between performance and cost. Our preference-conditioned routing mechanism captures the Pareto front of this optimization problem and adapts dynamically to user-specific preferences at inference time. The action-space aware routing policy generalizes to arbitrary sets of LLM models, demonstrated across various routing setups. We leverage a comprehensive evaluation framework that combines existing benchmarks and pairwise comparisons to assess routing effectiveness. Our efficient quizzing mechanism can characterize new models using only 20-50 carefully selected prompts, reducing the integration overhead by 90\% compared to full benchmark evaluation.

Experimental results across multiple benchmarks, including HELM, AlpacaEval, and OpenLLM Leaderboard, demonstrate that our method achieves up to 27\% improvement in cost-efficiency compared to existing routing approaches while maintaining comparable or better performance. The framework proves especially effective in real-world scenarios where both performance requirements and cost constraints vary across different applications and users. Our system enables organizations to optimize their LLM deployments by automatically selecting the most cost-effective model for each query while meeting performance requirements, particularly valuable for applications handling diverse query types or having varying performance demands.

\begin{figure}
    \centering
    \includegraphics[width=\linewidth]{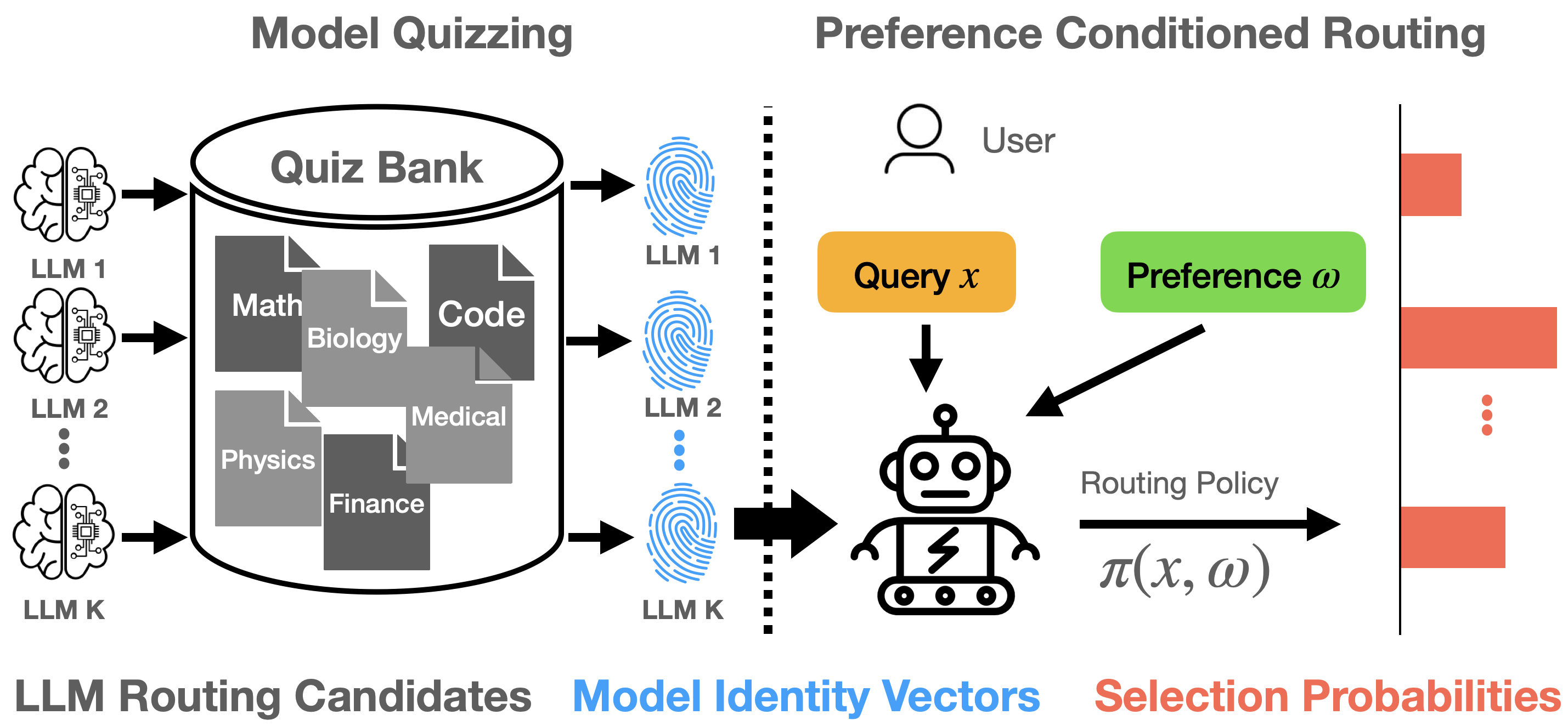}
    \vspace{-18pt}
    \caption{Overview of our preference-conditioned dynamic routing framework. Model quizzing (left) generates identity vectors capturing model capabilities, while routing policy (right) determines model selection based on user preferences and query.}
    \label{fig:llm_bandit}
    \vspace{-10pt}
\end{figure}

\vspace{-4pt}
\section{Method}
\subsection{Problem Formulation}\label{sec:problem_formulation}
Let $\mathcal{X}$ denote the space of all possible queries and $\{M_k\}_{k=1}^K$ be a finite set of $K$ large language models. Each model $M_k$ is characterized by its generation capabilities and an associated cost $c_k \in \mathbb{R}_+$. For any query $x \in \mathcal{X}$ and model $M_k$, we define $s(x, k) \in [0, 1]$ as a normalized score measuring the quality of $M_k$'s response to query $x$. This score can be obtained through various evaluation metrics (e.g., accuracy, F1-score) depending on the task domain.

We aim to develop a routing policy $\pi: \mathcal{X} \rightarrow \mathcal{P}(K)$, where $\mathcal{P}(K)$ denotes the probability simplex over $K$ models, that maps each query to a distribution over available models. When executing the policy, a model is sampled according to this distribution, i.e., $k \sim \pi(x)$. The routing decision results in a two-dimensional reward vector $\mathbf{r}(x, k) = [s(x, k), -c_k] \in \mathbb{R}^2$, capturing both the generation quality and the negative cost.

In the context of multi-objective optimization, we seek to maximize the expected reward vector:
\begin{equation*}
\mathbf{J}_\pi = \mathbb{E}_{x \sim p(x), k \sim \pi(x)}[\mathbf{r}(x, k)] = [\mathbb{E}_{x,\pi}[s(x, k)], -\mathbb{E}_{x,\pi}[c_k]],
\end{equation*}
where $p(x)$ denotes the underlying query distribution and $\mathbb{E}_{x,\pi}$ is shorthand for the expectation over both $x \sim p(x)$ and $k \sim \pi(x)$. Given two policies $\pi_1$ and $\pi_2$, we say $\pi_1$ dominates $\pi_2$ if $\mathbf{J}_{\pi_1} \geq \mathbf{J}_{\pi_2}$ elementwise and the inequality is strict in at least one dimension. A policy $\pi$ is Pareto optimal if it is not dominated by any other policy.

The set of all Pareto optimal policies forms the Pareto set $\Pi^*$, and their corresponding expected rewards $\{\mathbf{J}_\pi: \pi \in \Pi^*\}$ form the Pareto front. Due to the conflicting nature of performance and cost objectives, there typically exists no single policy that simultaneously maximizes both objectives. Instead, different policies in $\Pi^*$ represent different trade-offs between performance and cost.

To navigate this trade-off, we introduce a preference parameter $\boldsymbol{\omega} = [\omega_1, \omega_2] \in \mathbb{R}^2_+$ that specifies the relative importance of performance versus cost. This allows us to define a scalarized reward:
$
r_{\boldsymbol{\omega}}(x, k) = \boldsymbol{\omega}^\top \mathbf{r}(x, k) = \omega_1 s(x, k) - \omega_2 c_k.
$
For any fixed preference $\boldsymbol{\omega}$, the optimal policy $\pi_{\boldsymbol{\omega}}$ maximizes the expected scalarized reward:
\begin{equation}
\pi_{\boldsymbol{\omega}} = \argmax_\pi \mathbb{E}_{x \sim p(x), k \sim \pi(x)}[r_{\boldsymbol{\omega}}(x, k)].
\end{equation}
While the instantaneous reward $s(x,k)$ may be discrete (e.g., binary success/failure outcomes), the expected reward $\mathbb{E}_{x,\pi}[s(x,k)]$ is continuous in the policy parameters under mild regularity conditions on the policy class (see Theorem~\ref{thm:continuity} in Appendix). Specifically, when the policy $\pi$ is parameterized by continuous functions (e.g., neural networks with softmax outputs), the expected reward surface remains continuous despite discrete individual rewards. This ensures the existence of optimal policies $\pi_{\boldsymbol{\omega}}$ for each preference vector $\boldsymbol{\omega}$. Moreover, as $\boldsymbol{\omega}$ varies across $\mathbb{R}^2_+$, the corresponding optimal policies $\{\pi_{\boldsymbol{\omega}}: \boldsymbol{\omega} \in \mathbb{R}^2_+\}$ trace out the complete Pareto front of achievable performance-cost trade-offs \citep{yang2019generalized,basaklar2022pd}.

This formulation connects our problem to both multi-armed bandit \citep{katehakis1987multi,bouneffouf2019survey} and multi-objective optimization \citep{sharma2022comprehensive} literature. The routing policy must learn to select models (arms) based on query-specific contextual information, similar to contextual bandits. However, unlike traditional bandits that optimize a scalar reward, our setting involves vector-valued rewards and user-specified preferences, relating to multi-objective optimization theory. This combination presents unique challenges in policy learning and evaluation, which we address in subsequent sections.

\vspace{-4pt}
\subsection{Overall Framework}
Given the formulation above, our framework addresses two key challenges: (1) how to efficiently characterize each model's capabilities to enable informed routing decisions, and (2) how to learn a preference-conditioned policy that generalizes across different models and queries. We propose a two-component solution: a model quizzing component that generates identity vectors capturing model capabilities, and a preference-conditioned routing policy that determines selection probabilities. Figure~\ref{fig:llm_bandit} illustrates this framework.

\vspace{-4pt}
\subsection{Model Identity Vector}
To enable effective routing, we need a compact representation of each model's capabilities across different tasks and domains. Given a set of evaluation prompts $\mathcal{X} = \{x_n\}_{n=1}^N$ spanning various domains, we collect evaluation scores $\mathcal{Y}_k = \{y_{kn}\}_{n=1}^N$ for LLM $M_k$. Our goal is to learn a model identity vector $\mathbf{I}_k \in \mathbb{R}^d$ that predicts these evaluation scores.

We employ a variant of Item Response Theory (IRT) \citep{hambleton2013item} combined with deep neural networks. Unlike traditional IRT, we leverage pretrained prompt embeddings $\mathbf{e}_n$ rather than learning explicit prompt representations, enabling generalization to unseen prompts. The score prediction model $f(\mathbf{e}_n, \mathbf{I}_k)$ outputs the probability of model $M_k$ successfully handling prompt $x_n$.

For binary evaluation scores $\bar{y}_{kn}$, we optimize the binary cross-entropy loss:
\begin{equation}
\mathcal{L}_{\text{irt}} = \mathbb{E}[-\bar{y}_{kn}\log p_{kn} - (1-\bar{y}_{kn})\log(1-p_{kn})],
\end{equation}
where $p_{kn} = \text{sigmoid}(f(\mathbf{e}_n, \mathbf{I}_k))$. For non-binary scores, we employ a thresholding mechanism in Appendix~\ref{sec:appendix_binarize}.

We further incorporate pairwise model comparisons to enhance the identity vectors. Given responses from models $M_{k_1}$ and $M_{k_2}$ with annotations $z_n \in \{0,1\}$ indicating the winner, we introduce a secondary network $g$ that predicts winning probabilities:
\begin{equation}
\mathcal{L}_{\text{pair}} = \mathbb{E}[-z_n\log p_n - (1-z_n)\log(1-p_n)],
\end{equation}
where $p_n = \text{sigmoid}(g(\mathbf{e}_n, \mathbf{I}_{k_1}) - g(\mathbf{e}_n, \mathbf{I}_{k_2}))$.

To enhance generalization to unseen models, we employ variational inference, treating $\mathbf{I}_k$ as latent variables. This adds a KL-divergence term for regularization:
\begin{equation}
\mathcal{L}_{\text{KL}} = \mathbb{E}_k[D_{\text{KL}}(q(\mathbf{I}_k)\|p(\mathbf{I}_k))],
\end{equation}
where both prior $p(\mathbf{I}_k)$ and posterior $q(\mathbf{I}_k)$ are Gaussian distributions. Please see Appendix~\ref{sec:appendix_identity} for detailed derivations.

\vspace{-4pt}
\subsection{Preference-Conditioned Routing Policy}
Building on our problem formulation, the core challenge is to develop a routing policy that can (1) generalize across different sets of LLMs and (2) adapt to varying user preferences $\boldsymbol{\omega}$. A natural approach would be to directly estimate the evaluation scores $s(x,k)$ using our IRT model $f(\mathbf{e}, \mathbf{I}_k)$ and select models that maximize the scalarized reward $r_{\boldsymbol{\omega}}(x,k)$. However, this direct estimation approach faces several limitations. The predicted scores may be inaccurate for specific query-model pairs, the estimation provides no uncertainty quantification, and most importantly, the deterministic selection strategy cannot balance exploration and exploitation.

We propose to learn a stochastic policy $\pi_\theta$ that maps queries to routing decisions while incorporating both the available models and user preferences as conditioning information:
\begin{equation}
\pi_\theta(k' | x, \mathcal{C}_K, \boldsymbol{\omega}) \propto \exp(\mathbf{I}_{k'}^\top h(x, \mathcal{C}_K, \boldsymbol{\omega})).
\end{equation}
Our formulation introduces three key innovations to address the core challenges. First, we enable generalization across model sets through action-space awareness. The policy is explicitly conditioned on model identity vectors $\{\mathbf{I}_k\}_{k=1}^K$, making it aware of available actions. The dot-product structure between model identities and network outputs allows the policy to work with arbitrary sets of models - once we compute a model's identity vector, it can be immediately incorporated into routing decisions. 
Second, we enhance routing decisions by incorporating comprehensive context $\mathcal{C}_K = \{(\mathbf{I}_k, c_k, \hat{p}_k)\}_{k=1}^K$, which includes not only identity vectors but also costs $c_k$ and predicted scores $\hat{p}_k = \text{sigmoid}(f(x,\mathbf{I}_k))$. This context is processed through a permutation-invariant network $h(\cdot)$, enabling the policy to reason about relative strengths of different models for each specific query while maintaining consistency across different model orderings.
Third, we enable dynamic preference adaptation by directly conditioning the policy on $\boldsymbol{\omega}$. This allows the policy to adjust its routing strategy at inference time without retraining, efficiently exploring different performance-cost trade-offs based on user requirements.

We optimize the policy following standard multi-objective policy gradient algorithms \citep{xu2020prediction,shu2024learning}, where the gradient for updating the parameters $\theta$ is given by 
\begin{equation*}
    \nabla_{\theta}[\boldsymbol{\omega}^{T}\mathbf{J_{\pi_{\theta}}}] = \mathbb{E} \left[\boldsymbol{\omega}^{T}\mathbf{A}(x, k') \nabla_{\theta} \log \pi_{\theta}(k' \mid x, C_K, \boldsymbol{\omega}) \right],
\end{equation*}
where $\mathbf{A}(x,k')$ indicates the advantage function estimated from sampled trajectories. The corresponding value function $\mathbf{V}_{\pi_{\theta}}(x, C_K)$ outputs a vector of expected returns under the current policy $\pi_{\theta}$. The parameters of the value function are updated by a squared-error loss $\| \mathbf{V}_{\pi_{\theta}} - \mathbf{V}_{targ} \|^2$, where $\mathbf{V}_{targ}$ is the target value. Note the value function does not depend on the preference $\boldsymbol{\omega}$, which encourages shared values estimation across different user preferences. The vectorized value function is inspired by the core principles of multi-objective Q-learning algorithms \citep{yang2019generalized,basaklar2022pd}. This value network and policy gradient extension can be seamlessly integrated into most existing policy gradient methods. In our implementation, we adapt Proximal Policy Optimization (PPO) \citep{schulman2017proximal}, where the clipped surrogate objective is used to update policy parameters. Additionally, Generalized Advantage Estimation (GAE) \citep{schulman2015high} is employed to compute the advantage function $\mathbf{A}$ and target values $\mathbf{V}_{targ}$. For detailed derivations and implementation specifics, please refer to Appendix~\ref{sec:appendix_policy}.

A key advantage of our approach is its scalability. By leveraging model identity vectors and preference conditioning, the policy can seamlessly adapt to new models and varying user requirements without retraining from scratch. However, realizing these benefits requires careful consideration of training methodology. In the following sections, we explore the specific techniques that ensure effective generalization across models, queries, and preferences.

\vspace{-4pt}
\subsection{Training for Generalization}
While our policy architecture enables handling different models and preferences, realizing these capabilities requires careful training strategies. We identify three key generalization challenges: (1) handling arbitrary sets of models, (2) generalizing to unseen queries, and (3) maintaining consistent performance across preferences.

To handle arbitrary model sets, we employ two complementary strategies. First, we train the policy on dynamically sampled sets of models with varying sizes and capabilities. We leverage evaluation leaderboards like HELM \citep{liang2022holistic} that provide scores for diverse models, randomly selecting different combinations during training. This exposure to diverse model combinations forces the policy to learn generalizable routing strategies rather than memorizing specific model relationships. Second, we address the challenge of varying score and cost scales across different model combinations. For instance, comparing GPT-4 with Mixtral-8x7B yields different scales than comparing two open-source models. We handle this through reward normalization within each set:
\begin{equation}
    \bar{s}_k = s_k/\max(\{s_k\}_{k=1}^K), ~~~ \bar{c}_k = c_k/\max(\{c_k\}_{k=1}^K).
\end{equation}
This normalization ensures consistent reward scales regardless of the specific models involved, enabling stable optimization. Moreover, it maintains consistent interpretation of preference vectors - the same preference $\boldsymbol{\omega}$ will represent similar trade-offs across different model combinations.

For query generalization, we employ two techniques. First, we perform large-scale pretraining on pairwise model comparison datasets, such as Nectar \citep{starling2023} and Chatbot Arena \citep{zheng2023judging}. While these datasets feature diverse user queries that help learn generalizable routing behaviors, they only provide binary winning labels rather than model-specific evaluation scores. To leverage this data, we first obtain predicted scores from our IRT model, then calibrate them using Platt scaling \citep{platt1999probabilistic}:
$
    \bar{p}_k = \text{sigmoid}(\alpha f(x,\mathbf{I}_k) + \beta),
$
where $\alpha$ and $\beta$ are learned to align score predictions with human preferences (see Appendix~\ref{sec:appendix_pretraining}). The policy is pretrained to predict actions that maximize the calibrated reward:
$
    \hat{a} = \argmax_{k \in \{k_1, k_2\}} \boldsymbol{\omega}^T[\bar{p}_k, -c_k].
$
Second, we introduce an on-manifold mixup regularization during the subsequent reinforcement learning phase. When sampling queries from the replay buffer, we interpolate each prompt embedding with its nearest neighbor. This neighborhood-based interpolation ensures the mixed embeddings remain meaningful, helping the policy learn smoother decision boundaries (see Appendix~\ref{sec:appendix_mixup}).

For preference generalization, the key challenge is maintaining Pareto optimality while enabling efficient learning across different trade-offs. We leverage two complementary strategies. First, our decomposed value function $\mathbf{V}_{\pi_{\theta}}(x, C_K)$ estimates score and cost components independently. This decomposition enables value estimation sharing across preferences while maintaining separate tracking of objectives. Second, we train with dynamically sampled preferences $\boldsymbol{\omega} \sim U(\omega_{\min}, \omega_{\max})$, forcing the policy to learn consistent behaviors across different trade-offs. The preference range is chosen to cover practical trade-offs between quality and cost.

\vspace{-4pt}
\subsection{Cold Start for New Routing Candidates}
A key advantage of our framework is its ability to efficiently incorporate new LLMs without retraining the routing policy. When a new model $\tilde{M}$ is introduced, we only need to compute its identity vector $\tilde{\mathbf{I}}$ to enable routing. While this vector could be obtained through full benchmark evaluation, such an approach would be prohibitively expensive and time-consuming. We propose instead an efficient characterization method that requires evaluating only 20-50 carefully selected prompts, reducing the integration overhead by 90\% or more compared to full evaluation.

Our approach is based on the insight that not all evaluation prompts are equally informative for distinguishing model capabilities. Given existing prompts $\mathcal{X} = \{x_n\}_{n=1}^N$ and binary evaluation scores $\mathcal{Y}_k = \{\bar{y}_{kn}\}_{n=1}^N$ for existing models, we compute a discrimination score for each prompt:
\begin{equation}
    \psi_n = \mathbb{E}_k [-\bar{y}_{kn}\log p_{kn} - (1-\bar{y}_{kn})\log(1-p_{kn})]
\end{equation}
where $p_{kn} = \text{sigmoid}(f(x_n, \mathbf{I}_k))$ indicates IRT model's prediction. The score $\psi_n$ measures the average prediction error across models for prompt $x_n$. A high $\psi_n$ indicates that our IRT model struggles to accurately predict model performance on this prompt, often because models with similar capabilities exhibit inconsistent performance. Conversely, a low $\psi_n$ suggests model performance is highly predictable - either consistently successful or unsuccessful across models with similar capabilities.

Using these discrimination scores, we select a representative subset of prompts $\tilde{\mathcal{X}}$ through stratified sampling. By sampling from different strata of $\psi_n$ values, we ensure our evaluation set covers prompts with varying discriminative power. Given the new model's evaluation scores $\tilde{Y}$ on these selected prompts, we compute its identity vector by:
\begin{equation}\label{eq:id_optim}
\tilde{\mathbf{I}}=\argmin_{\mathbf{I}} [\mathbb{E}_{\tilde{\mathcal{X}}}[-\bar{y}\log p-(1-\bar{y})\log(1-p)]+\KL(q\|p)],
\end{equation}
where $p = \text{sigmoid}(f(x, \mathbf{I}))$. The KL term acts as a regularizer, encouraging the identity vector to remain close to our prior distribution over model capabilities. Importantly, this optimization updates only the identity vector $\tilde{\mathbf{I}}$ while keeping the IRT model fixed. Once computed, the identity vector $\tilde{\mathbf{I}}$ can be immediately used by our routing policy without any additional training or fine-tuning. This is enabled by our dot-product architecture that naturally extends to new models. As shown in Fig.~\ref{fig:new_model}, routing performance with these efficiently computed identity vectors closely matches that of vectors computed using full evaluation.

The combination of discriminative prompt selection and efficient identity vector computation provides a practical solution for maintaining an up-to-date routing system in the rapidly evolving LLM landscape. Implementation details and additional analysis can be found in Appendix~\ref{sec:appendix_stratified_sampling}.

\vspace{-4pt}
\section{Related Works}
\textbf{LLM Ensemble, Cascade and Routing}\quad
As the number of LLMs grows, there is increasing interest in combining them to optimize performance and balance costs. LLM ensemble methods improve response quality by aggregating outputs from multiple LLMs but incur high computational costs since they require running inference on multiple models \citep{jiang2023llm,wang2023fusing,lu2024blending}. LLM cascading reduces costs by invoking LLMs sequentially, starting with the least expensive model and progressing to more costly ones until a satisfactory response \citep{chen2023frugalgpt,madaan2023automix,ramirez2024optimising}. While effective in reducing costs, cascading still requires multiple inferences, especially for complex queries, and often depends on an additional model to assess the response quality. 

In contrast, LLM routing sends queries directly to the most appropriate model, requiring only a single inference and thus offering a more cost-efficient solution. Typical routing methods rely on performance prediction models to guide the selection of LLM. These methods either predict downstream evaluation or reward scores for a given query \citep{shnitzer2023large,lu2023routing,hari2023tryage,vsakota2024fly}, or estimate win rates between pairs of models \citep{ding2024hybrid,ong2024routellm}. The chosen LLM is then selected based on predicted performance and any additional constraints, such as cost or latency. 

The most relevant work to ours is MetaLLM \citep{nguyen2024metallm}, which also frames the routing task as a multi-armed bandit problem. However, MetaLLM optimizes a scalarized reward and operates on a fixed set of LLMs, limiting the learned policy to specific user preferences and a predefined set of models. Our approach, by contrast, generalizes to varied user preferences and dynamically adapts to new LLMs added to the system, ensuring broader applicability and greater flexibility.

\textbf{Multi-objective Reinforcement Learning}~~
Multi-objective RL seeks to optimize multiple, often conflicting reward signals within a Markov decision process, resulting in a set of Pareto-optimal policies known as the Pareto set rather than a single optimal policy. Traditional algorithms typically aim to approximate this Pareto set by searching for a finite number of policies \citep{van2014multi,parisi2014policy,xu2020prediction}. However, these methods face the curse of dimensionality, where the number of policies needed to accurately approximate the Pareto set grows exponentially with the number of objectives. To address this, recent approaches have proposed using a single deep neural network conditioned on preferences to represent the entire Pareto set \citep{yang2019generalized,abels2019dynamic,basaklar2022pd}. Another approach involves using hypernetworks \citep{chauhan2023brief}, which map user preferences to the parameters of the policy network \citep{shu2024learning}. Our routing policy aligns with the conditional neural network framework, where a single model is conditioned on user preferences to adapt to different user requirements. We further tailor this conditional architecture specifically for routing LLMs, allowing for efficient decision-making across a diverse and expanding set of models.

\textbf{Generalization in Reinforcement Learning}\quad
Generalizing RL policies to new tasks, particularly zero-shot RL, focuses on enabling policies to handle unseen tasks without retraining \citep{korkmaz2024survey}. Existing approaches either maximize worst-case performance through adversarial training \citep{moos2022robust,dong2023robust}, compute task representations from exploration data \citep{touati2021learning,agarwal2021contrastive,benjamins2022contextualize,ingebrand2024zero}, or leverage in-context learning with transformers \citep{melo2022transformers,brohan2022rt}. Our routing policy adopts the representation-based approach, where the task representation is explicitly provided as a set of LLMs and their associated costs, similar to \citet{jain2020generalization}'s work on action space generalization. For observation distribution generalization, we employ a simple regularization technique to encourage smoothness across prompt distributions, building on established techniques in RL generalization literature \citep{cobbe2019quantifying,zhang2021generalization}. A detailed discussion of related approaches is provided in Appendix~\ref{sec:appendxi_related_works}.

\vspace{-4pt}
\section{Experiments}
\vspace{-2pt}
We evaluate our routing policy on 5 popular LLM benchmarks, including HELM-Lite \citep{liang2022holistic} HELM-MMLU \citep{liang2022holistic}, HuggingFace OpenLLM Leaderboard \citep{open-llm-leaderboard}, HuggingFace OpenLLM Leaderboard v2 \citep{open-llm-leaderboard-v2}, and AlpaceEval 2.0 \citep{li2023alpacaeval}. We divide the prompts in each leaderboard into training and test splits. The training split is used to train the routing policy, and the test split is reserved to evaluate the routing performance. The routing policy is first pretrained on pairwise comparison datasets, including Chatbot Arena \citep{zheng2023judging}, Nectar \citep{starling2023}, and a synthetic dataset from RouteLLM \citep{ong2024routellm}. We train the IRT model on the same pairwise datasets and the training splits of all leaderboards, and we use this IRT model across all evaluations. For cost estimation, we approximate the model invocation costs based on processing and generating 1M tokens each. We evaluate our routing policy across various LLM candidates. Please refer to Appendix~\ref{sec:appendix_experiment} for further details on cost estimation and experimental setup.

Following RouteLLM \citep{ong2024routellm}, we evaluate our approach in a scenario with two LLM candidates, GPT-4 and Mixtral-8x7B, and compare it to RouteLLM. It is important to note that while RouteLLM is specifically trained for this two-model configuration, our routing model is designed to handle arbitrary sets of LLM candidates. To further test generalization, we evaluate two additional LLM configurations for each dataset. In these multi-LLM settings, RouteLLM is not applicable, as it is restricted to two candidates. For RouteLLM, we adjust the routing preference by specifying different thresholds, whereas our routing models can directly take preferences as inputs. When evaluating scenarios with only two LLM candidates, we also compare against a random baseline, where the model is selected randomly, and the preference is adjusted according to the selection probability. However, in scenarios with more than two LLMs, adjusting the selection probability becomes non-trivial, so we omit the random baseline in these cases. We also compare against a baseline that uses the predicted scores $\hat{p}_k$ to compute utility (denoted as \textit{Predictor}). Another baseline involves training a PPO routing policy to optimize the scalarized reward $r_{\boldsymbol{\omega}}$, but this requires separate policies for each LLM set and preference. Lastly, we introduce an oracle policy that selects the LLM based on the actual evaluation scores $s(x, k)$.

\begin{figure*}
    \begin{minipage}{0.02\linewidth}
        \centering
        \rotatebox{90}{\textbf{AlpacaEval 2.0}}
    \end{minipage}
    \begin{minipage}{0.97\linewidth}
        \centering
        \subfigure[GPT4/Mixtral-8x7B]{\includegraphics[width=0.28\linewidth]{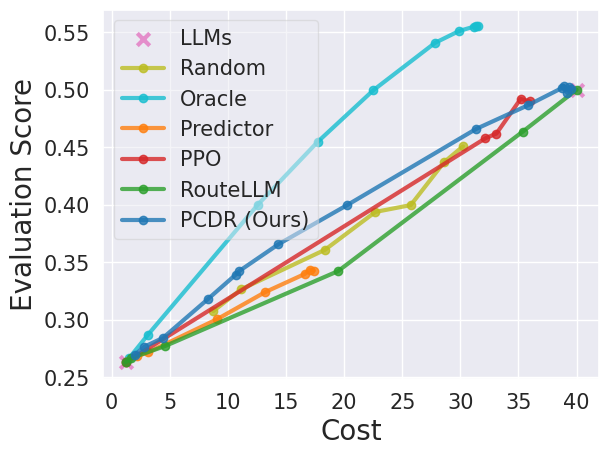}}
        \subfigure[GPT Family]{\includegraphics[width=0.28\linewidth]{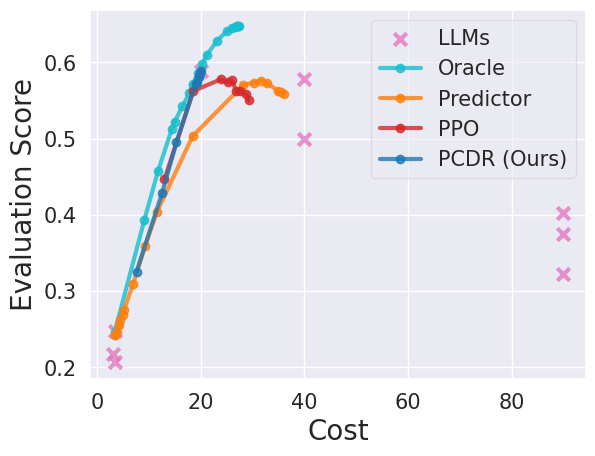}}
        \subfigure[Claude Family]{\includegraphics[width=0.28\linewidth]{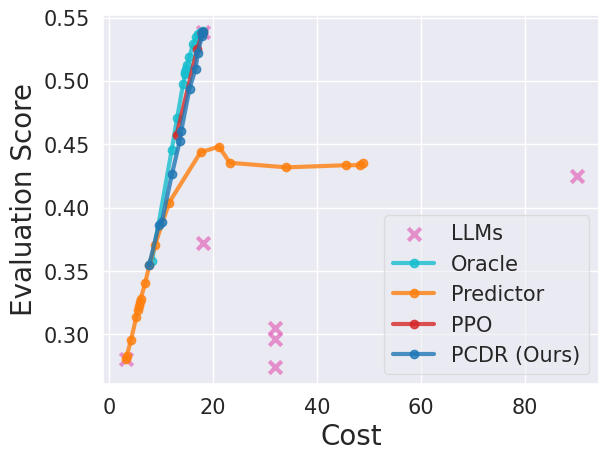}}
    \end{minipage}
    \setcounter{subfigure}{0}
    \begin{minipage}{0.02\textwidth}
        \centering
        \rotatebox{90}{\textbf{HELM-MMLU}}
    \end{minipage}
    \begin{minipage}{0.97\textwidth}
        \centering
        \subfigure[GPT4/Mixtral-8x7B]{\includegraphics[width=0.28\linewidth]{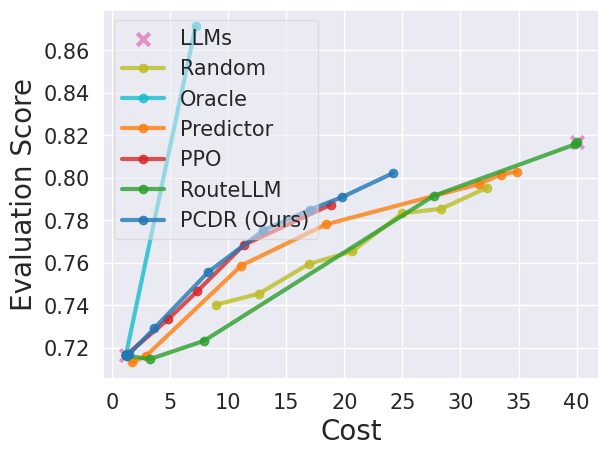}}
        \subfigure[Mistral Family]{\includegraphics[width=0.28\linewidth]{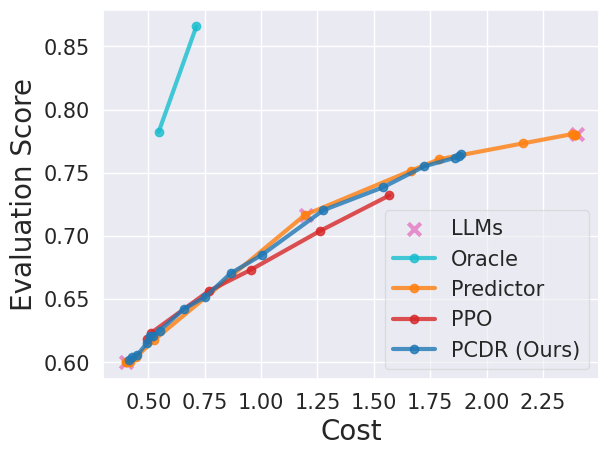}}
        \subfigure[GPT Family]{\includegraphics[width=0.28\linewidth]{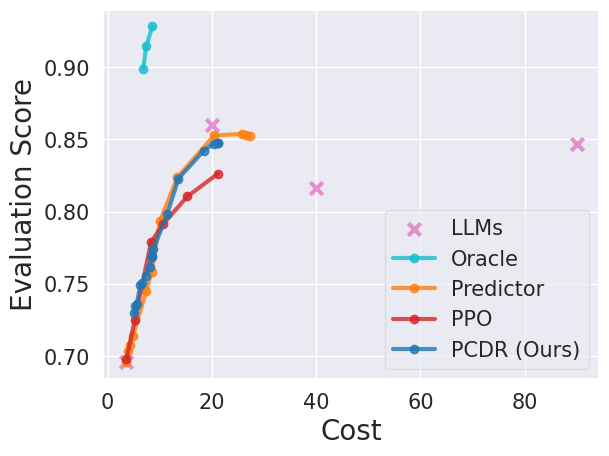}}
    \end{minipage}
    \setcounter{subfigure}{0}
    \begin{minipage}{0.02\textwidth}
        \centering
        \rotatebox{90}{\textbf{HELM-Lite}}
    \end{minipage}
    \begin{minipage}{0.97\textwidth}
        \centering
        \subfigure[GPT4/Mixtral-8x7B]{\includegraphics[width=0.28\linewidth]{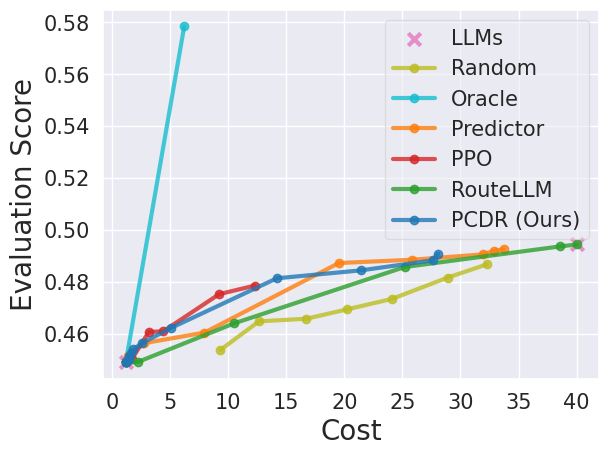}}
        \subfigure[Mistral Family]{\includegraphics[width=0.28\linewidth]{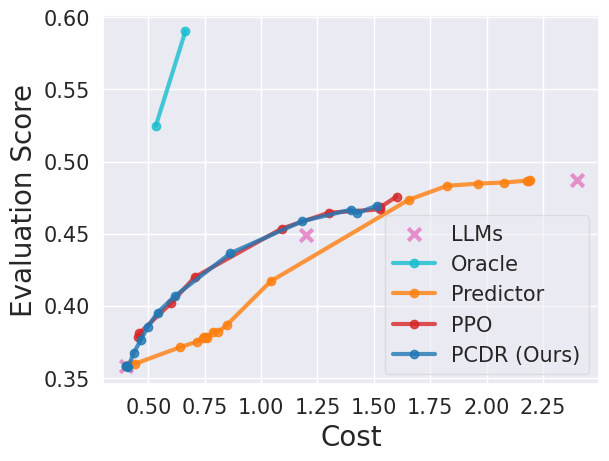}}
        \subfigure[GPT Family]{\includegraphics[width=0.28\linewidth]{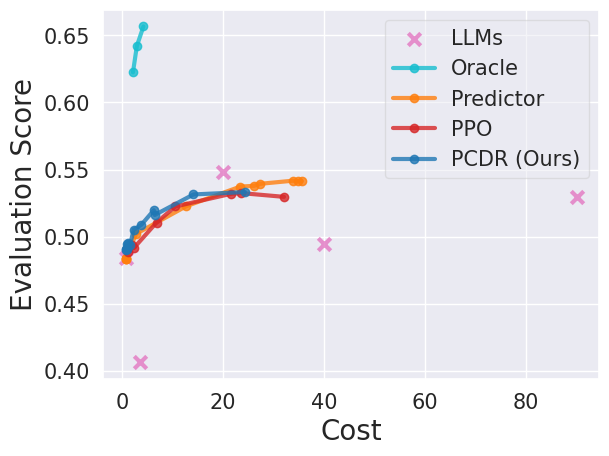}}
    \end{minipage}
    \setcounter{subfigure}{0}
    \begin{minipage}{0.02\textwidth}
        \centering
        \rotatebox{90}{\textbf{OpenLLM}}
    \end{minipage}
    \begin{minipage}{0.97\textwidth}
        \centering
        \subfigure[Yi1.5 Family]{\includegraphics[width=0.28\linewidth]{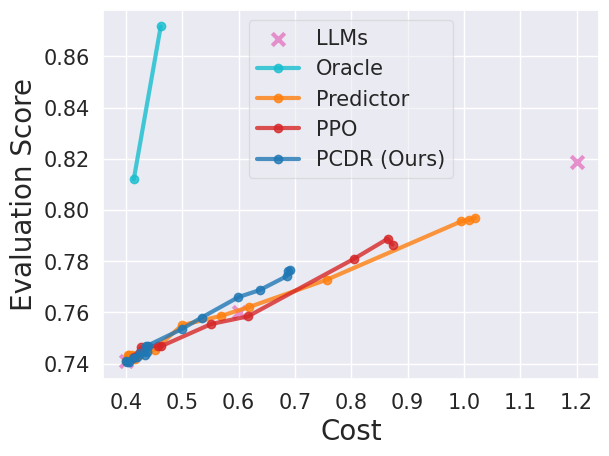}}
        \subfigure[Mistral Family]{\includegraphics[width=0.28\linewidth]{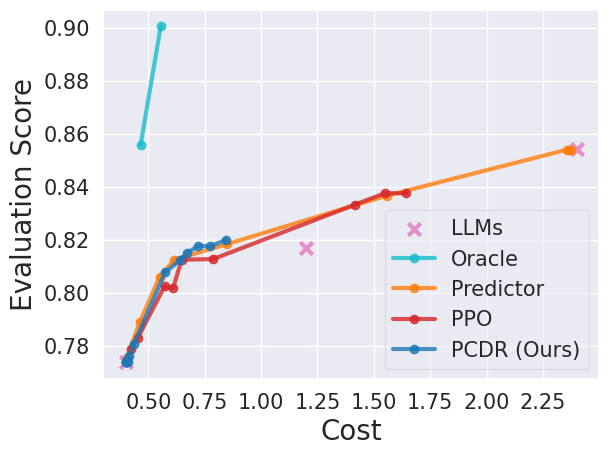}}
        \subfigure[LLaMA3 Family]{\includegraphics[width=0.28\linewidth]{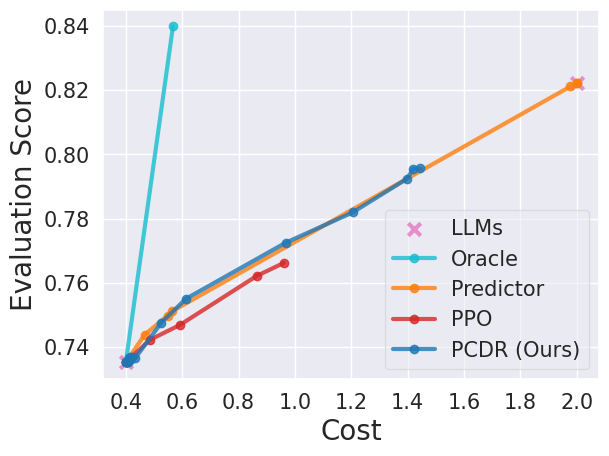}}
    \end{minipage}
    \setcounter{subfigure}{0}
    \begin{minipage}{0.02\textwidth}
        \centering
        \rotatebox{90}{\textbf{OpenLLMv2}}
    \end{minipage}
    \begin{minipage}{0.97\textwidth}
        \centering
        \subfigure[Yi1.5 Family]{\includegraphics[width=0.28\linewidth]{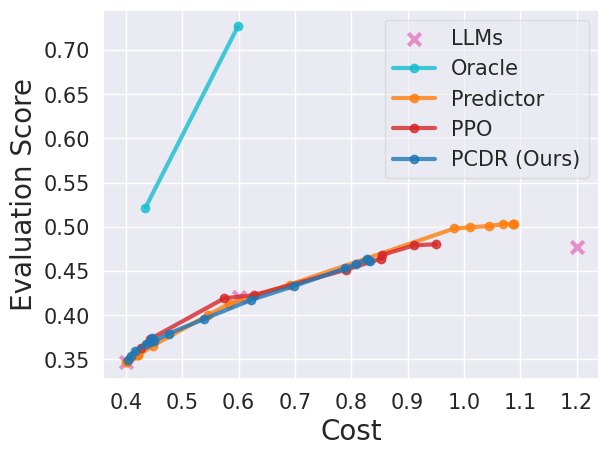}}
        \subfigure[Qwen2 Family]{\includegraphics[width=0.28\linewidth]{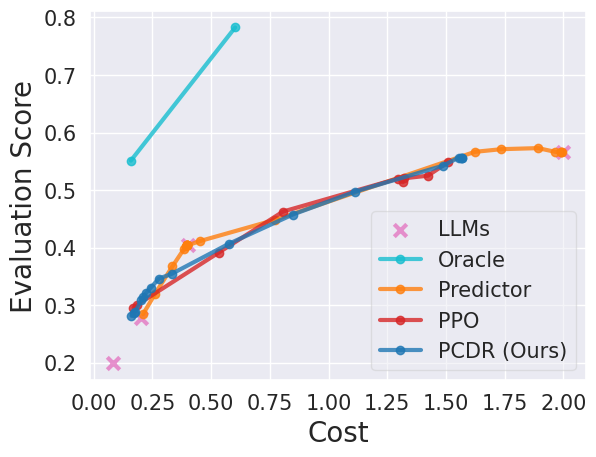}}
        \subfigure[LLaMA3 Family]{\includegraphics[width=0.28\linewidth]{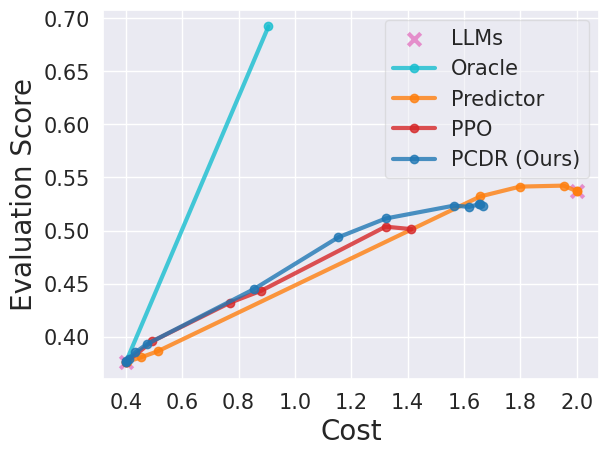}}
    \end{minipage}
    \caption{Evaluate the routing performance across 5 datasets and various sets of LLM candidates.}
    \label{fig:results}
\end{figure*}

\textbf{Results}\quad
Figure~\ref{fig:results} demonstrates our method's routing performance across five major LLM evaluation benchmarks with various model combinations. The results reveal several key advantages of our approach:
First, our Predictor baseline consistently outperforms RouteLLM, validating the effectiveness of our model identity vector and score prediction framework. Second, our preference-conditioned routing policy further improves upon the Predictor baseline, particularly in challenging scenarios where score prediction is less reliable. This is especially evident in AlpacaEval 2.0 (c) and HELM-Lite (b), where the routing policy learns to compensate for prediction uncertainties by incorporating broader context about model capabilities and costs. When compared to RouteLLM, our policy demonstrates substantial cost savings - on AlpacaEval 2.0 with the GPT4/Mixtral-8x7B configuration, our approach achieves 46.35\% accuracy at \$31 cost compared to RouteLLM's \$35, representing an 11\% cost reduction. On MMLU, the improvement is even more significant, reducing costs from \$33 to \$24 (27\% reduction) while maintaining 80\% accuracy. Third, our single routing policy achieves comparable or better performance than separately trained PPO policies across all datasets and LLM configurations. This is a crucial advantage, as each PPO baseline requires specific training for its fixed set of models and preference settings, while our approach generalizes across arbitrary model combinations and preferences without retraining. This demonstration of robust generalization is particularly important for practical deployments where model sets and requirements frequently change. While these results demonstrate significant improvements over existing methods, the performance gap between all routing policies and the Oracle baseline indicates potential for further optimization. This gap suggests opportunities for future work in improving both prediction accuracy and routing strategy.

\begin{figure}
    \centering
    \subfigure[Cohere]{\includegraphics[width=0.49\linewidth]{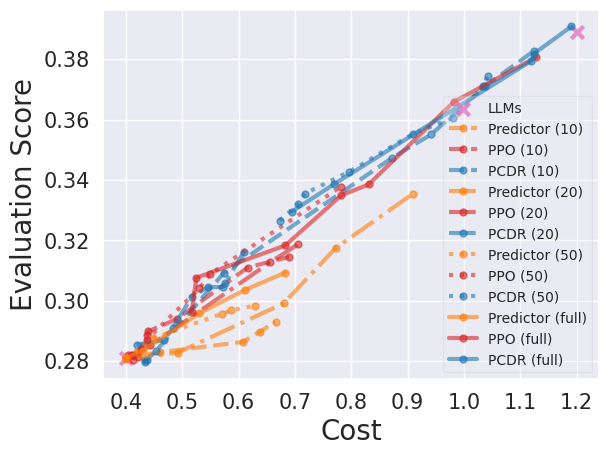}}
    \subfigure[Qwen2.5]{\includegraphics[width=0.49\linewidth]{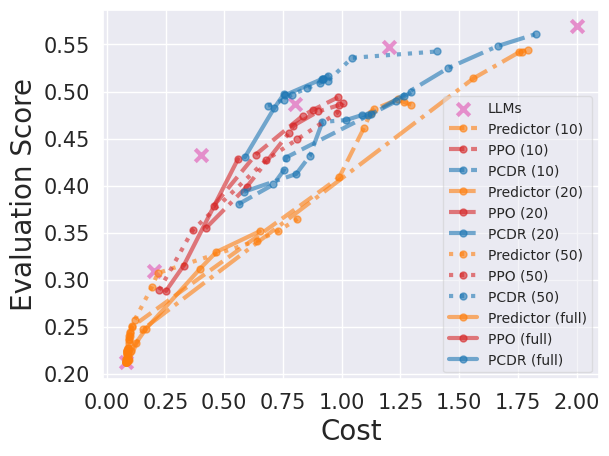}}
    \vspace{-10pt}
    \caption{Evaluate routing performance on two sets of new models. the identity vectors are obtained using 10, 20 or 50 selected prompts, respectively.}
    \label{fig:new_model}
    \vspace{-14pt}
\end{figure}

\textbf{Cold Start for New Routing Candidates}\quad
To simulate the scenario where new models are introduced into the routing system, we select several unseen models from the HuggingFace OpenLLM v2 benchmark. These models are not used for training either the IRT model or the routing policy. For detailed evaluation settings, please refer to  Appendix~\ref{sec:appendix_new_model}. The identity vectors for these models are obtained by optimize \eqref{eq:id_optim} over a selected subset of prompts from the OpenLLMv2 benchmark. We explore different evaluation budgets, selecting $10$, $20$ or $50$ prompts to obtain the evaluation scores for these newly added models. The \textit{Predictor} baseline utilizes the learned identity vectors to predict the evaluation scores, while the \textit{PPO} baseline trains the routing policy using the same set of selected prompts. For our preference conditioned routing policy, we directly plug the identity vectors into the routing policy trained on OpenLLMv2 (as shown in the last row of Figure~\ref{fig:results}), without further tuning on these newly added models. Figure~\ref{fig:new_model} presents the evaluation results. Overall, our routing policy outperforms the \textit{Predictor} baseline and performs comparably to the \textit{PPO} policy, despite the latter being specifically trained on the new models. Additionally, our approach maintains effectiveness even with very limited evaluation data - using just 50 prompts achieves performance nearly matching that of identity vectors computed from the full set.

\textbf{Computational Overhead}\quad
The routing overhead is minimal compared to model inference time. Our policy requires approximately 5ms per routing decision on a single GPU, negligible compared to typical LLM inference times (100ms-1s). Memory requirements are also modest: the identity vectors and routing policy together require less than 100MB of GPU memory.

\begin{wrapfigure}{r}{0.5\linewidth}
    \centering
    \vspace{-15pt}
    \includegraphics[width=\linewidth]{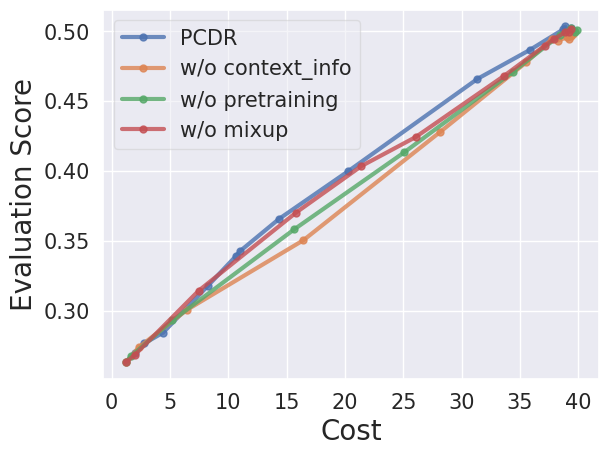}
    \vspace{-18pt}
    \caption{Ablation studies on the routing policy components.}
    \label{fig:ablation}
    \vspace{-8pt}
\end{wrapfigure}
\textbf{Ablation Studies}
Our routing policy consists of a supervised pretraining stage followed by a RL training stage. During training, we incorporate on-manifold mixup regularization to improve generalization to unseen prompts. Additionally, our policy leverages the predicted scores $\hat{p}_k$ as contextual information. In this section, we perform ablation studies to assess the contributions of these components. Figure~\ref{fig:ablation} presents the results when each component is removed. The results indicate that the context information, pretraining stage, and mixup regularization all contribute to learning a more effective routing policy.

\vspace{-4pt}
\section{Conclusion}
\vspace{-2pt}
In this work, we present a novel preference-conditioned dynamic routing framework for large language models that addresses three key challenges in LLM deployment: balancing performance and cost, adapting to user preferences, and incorporating new models efficiently. We formulate LLM routing as a multi-objective optimization problem and develop a preference-conditioned policy that dynamically adapts to user requirements at inference time. Our approach introduces a model identity vector framework that enables efficient integration of new LLMs without policy retraining, reducing adaptation time from hours to minutes. Through comprehensive experiments on five major benchmarks, we demonstrate significant improvements over existing methods, achieving up to 27\% cost reduction while maintaining comparable performance.

Our results highlight the potential of intelligent routing systems in making LLM deployments more efficient and adaptable. However, several promising directions remain for future research. While our current approach operates in an offline setting with pre-computed evaluation scores, extending to online learning could improve policy robustness through real-time feedback. This would enable continuous adaptation to changing user needs and model performance patterns. Our framework currently assumes fixed costs per model, but real-world costs vary with input length and computation requirements. Developing adaptive cost models that account for query-specific characteristics could enable more precise optimization of the performance-cost trade-off.

Future work could also expand the routing capability to leverage external tools and API calls that many modern LLMs support. This could include incorporating tool use, online search results, and other augmentations into the routing decision process. Additionally, while our preference-based approach offers flexibility, expressing trade-offs through numerical parameters may not be intuitive for all users. Developing more natural interfaces for preference specification and automated preference learning from user feedback could improve usability.

As the LLM ecosystem continues to evolve with new models and capabilities, efficient routing systems will become increasingly critical for practical applications. Our framework provides a foundation for building more sophisticated, adaptive, and user-friendly LLM routing systems that can meet the diverse needs of real-world deployments.




\section*{Impact Statement}
Our work on efficient LLM routing has several potential societal implications. On the positive side, by enabling more cost-efficient use of LLMs, our approach could help democratize access to advanced AI capabilities, allowing organizations with limited resources to leverage these technologies more effectively. The ability to balance performance and cost dynamically could make AI applications more sustainable and economically viable for a broader range of users.

However, this work also raises important considerations. By making LLM deployments more efficient, we could accelerate the adoption of these technologies, potentially exacerbating existing concerns about AI's impact on privacy, misinformation, and labor markets. Additionally, while our routing system aims to optimize resource allocation, it could inadvertently reinforce biases present in the underlying models if not carefully monitored.

To address these concerns, we emphasize that our framework is designed to be transparent in its decision-making process and configurable to align with organizational policies and ethical guidelines. We encourage users of this technology to carefully consider their specific use cases and implement appropriate safeguards, particularly when deploying in sensitive domains.

\bibliography{main}
\bibliographystyle{icml2025}

\newpage
\appendix
\onecolumn

\counterwithin{figure}{section}
\renewcommand\thefigure{\thesection.\arabic{figure}}
\counterwithin{table}{section}
\renewcommand\thetable{\thesection.\arabic{table}}
\counterwithin{equation}{section}
\renewcommand\theequation{\thesection.\arabic{equation}}

\section{Theoretical Analysis}

\subsection{Continuity and Existence of Optimal Policies}

Consider a policy $\pi_\theta$ parameterized by $\theta \in \Theta$, where $\Theta$ is a compact subset of $\mathbb{R}^d$. Let $s(x,k) \in \{0,1\}$ be a binary reward function and $p(x)$ be the query distribution.

\begin{theorem}
\label{thm:continuity}
If the policy $\pi_\theta(k|x)$ is continuous in $\theta$ for all $x$ and $k$, then the expected reward $J(\theta) = \mathbb{E}_{x \sim p(x), k \sim \pi_\theta(x)}[s(x,k)]$ is continuous in $\theta$.
\end{theorem}

\begin{proof}
For any $\theta, \theta' \in \Theta$:
\begin{align*}
|J(\theta) - J(\theta')| &= \left|\int_\mathcal{X} \sum_k s(x,k)(\pi_\theta(k|x) - \pi_{\theta'}(k|x))p(x)dx\right| \\
&\leq \int_\mathcal{X} \sum_k |s(x,k)||\pi_\theta(k|x) - \pi_{\theta'}(k|x)|p(x)dx \\
&\leq \int_\mathcal{X} \sum_k |\pi_\theta(k|x) - \pi_{\theta'}(k|x)|p(x)dx
\end{align*}
Since $\pi_\theta(k|x)$ is continuous in $\theta$, for any $\epsilon > 0$, there exists $\delta > 0$ such that $\|\theta - \theta'\| < \delta$ implies $|\pi_\theta(k|x) - \pi_{\theta'}(k|x)| < \epsilon/K$ for all $k$ and $x$. Therefore, $|J(\theta) - J(\theta')| < \epsilon$ when $\|\theta - \theta'\| < \delta$, proving continuity.
\end{proof}

\begin{corollary}
For any preference vector $\boldsymbol{\omega}$, there exists an optimal policy $\pi_{\boldsymbol{\omega}}$ that maximizes the expected scalarized reward.
\end{corollary}

This follows from the extreme value theorem, as we are maximizing a continuous function over a compact set. For the relationship between the preference vectors and the Pareto front, we refer readers to \citet{yang2019generalized} who provide a detailed analysis in the context of multi-objective reinforcement learning.

\section{Method}

\subsection{Model Identity Vector}\label{sec:appendix_identity}
We learn the model identity vector $\mathbf{I}_k$ following a variational variant of the IRT model. Given evaluation scores $Y_k = \{y_{kn}\}_{n=1}^N$ for model $M_k$ on a set of prompts $X = \{x_n\}_{n=1}^N$, we maximize the following variational lower bound of the log-likelihood:
\begin{equation}
\begin{aligned}
    \log p(y_{kn} \mid x_n) &= \log \int p(y_{kn}, \mathbf{I}_k \mid x_n) d \mathbf{I}_k\\
    & \geq \mathbb{E}_{q(\mathbf{I}_k)} \left[ \log p(y_{kn} \mid x_n, \mathbf{I}_k) \right] - \KL(q(\mathbf{I}_k) \| p(\mathbf{I}_k)).
\end{aligned}
\end{equation}
Here, the model embedding $\mathbf{I}_k$ is treated as a latent variable, with the posterior and prior distributions over $\mathbf{I}_k$ denoted by $q(\mathbf{I}_k)$ and $p(\mathbf{I}_k)$, respectively. In practice, both distributions are modeled as Gaussians, with the posterior $q(\mathbf{I}_k) = \mathcal{N}(\mathbf{\mu}_k, \Sigma_k)$ and the prior $p(\mathbf{I}_k) = \mathcal{N}(\mathbf{0}, \mathbf{I})$. The posterior mean $\mathbf{\mu}_k$ and variance $\Sigma_k$ are represented as embedding vectors of dimension $d$, with the variance assumed to be diagonal. The predictive distribution $p(y_{kn} \mid x_n, \mathbf{I}_k)$ is implemented as a neural network that concatenates of prompt and model embeddings as input and outputs the score prediction logits. During training, the loss is computed over the entire evaluation benchmarks, involving multiple prompts and models, i.e., $-\mathbb{E}_{x, k} \log p(y_{kn} \mid x_n)$.

\subsubsection{Training with Real-valued Evaluation Scores}\label{sec:appendix_binarize}
Certain evaluation datasets produce real-valued evaluation scores, such as F1 and RougeL. In order to unify the training procedure, we propose to binarize the real-valued scores. Specifically, given a set of real-valued scores $Y=\{y_{n}\}_{n=1}^N$, where $y_{n} \in [0, 1]$, we find an optimal threshold $\eta^*$ so that the average performance across instances are close to the original scores, that is
\begin{equation}
    \eta^* = \argmin_{\eta} \left(\frac{1}{N}\sum_{n=1}^N \mathbb{I}(y_n > \eta) - \frac{1}{N}\sum_{n=1}^N y_n \right)^2,
\end{equation}
where $\mathbb{I}(y_n > \eta)$ is the indicator function, which equals to 1 only when the condition $y_n > \eta$ is true. Therefore, the binarized evaluation scores are derived as $\bar{Y} = \{\mathbb{I}(y_n > \eta^*)\}_{n=1}^N$. 

\subsection{Preference Conditioned Routing Policy}\label{sec:appendix_policy}
In the main text, we derived the routing policy as 
\begin{equation*}
    \pi_{\theta}(k' \mid x, \{(\mathbf{I}_k, c_k, \hat{p}_k)\}_{k=1}^K, \boldsymbol{\omega}) \propto I_{k'}^{T} h(x, \{(\mathbf{I}_k, c_k, \hat{p}_k)\}_{k=1}^K, \boldsymbol{\omega}),
\end{equation*}
where $h(\cdot)$ is a neural network that is permutation invariant to the set $\{(\mathbf{I}_k, c_k, \hat{p}_k)\}_{k=1}^K$. We achieve the permutation invariance by using a permutation invariant embeddings of the set, implemented via the SetTransformer architecture\citep{lee2019set}. The prompt $x$ is encoded using pretrained prompt embeddings. The preference vector $\boldsymbol{\omega}$ is projected through a linear layer for integration into the routing policy. The neural network then concatenates the embeddings and passes them through several linear layers, resulting in a vector representation in $\mathbb{R}^d$. The inner product between $h(\cdot)$ and each model embedding $I_{k'}$ determines which model to select based on the policy. Specifically, the routing probability for selecting model $M_{k'}$ follows the softmax distribution:
\begin{equation}
    \pi_{\theta}(k' \mid x, \{(\mathbf{I}_k, c_k, \hat{p}_k)\}_{k=1}^{K}, \boldsymbol{\omega}) = \frac{\exp \left( I_{k'}^{T}h(x, \{(\mathbf{I}_k, c_k, \hat{p}_k)\}_{k=1}^{K}, \boldsymbol{\omega}) \right)}{\sum_{k''=1}^{K} \exp \left( I_{k''}^{T}h(x, \{(\mathbf{I}_k, c_k, \hat{p}_k)\}_{k=1}^{K}, \boldsymbol{\omega}) \right)}.
\end{equation}

We train the routing policy following the multi-objective PPO algorithm, where the gradient for updating the policy parameters $\theta$ is given by 
\begin{equation*}
    \nabla_{\theta}[\boldsymbol{\omega}^{T}\mathbf{J_{\pi_{\theta}}}] = \mathbb{E}_{x,k'} \left[\boldsymbol{\omega}^{T}\mathbf{A}(x, k') \nabla_{\theta} \log \pi_{\theta}(k' \mid x, \{(\mathbf{I}_k, c_k, \hat{p}_k)\}_{k=1}^K, w) \right],
\end{equation*}
where $\mathbf{A}(x,k')$ indicates the advantage function estimated via GAE \citep{schulman2015high}. The PPO algorithm also requires a value estimation to reduce the gradient variance. Following multi-objective RL literature \citep{xu2020prediction,shu2024learning}, we define a value network $\mathbf{V}_{\pi_{\theta}}(x, \{(\mathbf{I}_k, c_k, \hat{p}_k)\}_{k=1}^{K})$ that outputs a vector of expected returns under the current policy $\pi_{\theta}$. The value estimation is not conditioned on the preference, therefore, it can be shared across different user preferences. We train the values network by optimizing a MSE loss $\| \mathbf{V}_{\pi_{\theta}} - \mathbf{V}_{targ} \|^2$, where $\mathbf{V}_{targ}$ indicates the target values estimated via GAE.

\subsection{Generalization of the Routing Policy}\label{sec:appendix_generalization}
In this section, we discuss the training procedure of the dynamic routing policy, which is designed to enhance the generalizability of the policy to various scenarios.

\subsubsection{Supervised Pretraining}\label{sec:appendix_pretraining}
The supervised pretraining stage leverages diverse prompts from pairwise comparison datasets to enhance generalization to unseen prompts. Given a pairwise comparison dataset $\mathcal{V}$, where each example consists of a prompt $x_n$, a pair of models $M_{k1}$ and $M_{k2}$, and a winning label $z_n \in \{0, 1\}$, we first train a logistic regression model to calibrate the predicted evaluation scores, $\hat{p} = \text{sigmoid}(f(x, \mathbf{I}_k))$, using the winning label $z_n$. Specifically, the logical regression model predicts the wining probability as $p(z_n = 1) = \text{sigmoid}(\alpha (f(x_n, I_{k1}) - f(x_n, I_{k2})) + \beta)$, where $\alpha$ and $\beta$ are learnable parameters. After training, the calibrated evaluation scores are given by $\bar{p} = \text{sigmoid}(\alpha f(x, \mathbf{I}_k) + \beta)$. The calibration follows the well-known Platt scaling \citep{platt1999probabilistic} algorithm, which refines the evaluation scores using human-labeled winning labels to produce more accurate predictions. 

With the calibrated evaluation scores $\bar{p}$ on a prompt $x$ and a user preference vector $\boldsymbol{\omega}$, the routing action is determined by $\hat{a} = \argmax_{k \in \{k1, k2\}} \boldsymbol{\omega}^T[\bar{p}_k, -c_k]$. We then pretrain the routing policy in a supervised manner using the following negative log-likelihood loss:
\begin{equation}
    \mathcal{L}_{pretrain} = - \log \pi(\hat{a} \mid x, \{(\mathbf{I}_k, c_k, \hat{p}_k)\}_{k \in \{k1, k2\}}, \boldsymbol{\omega}).
\end{equation}
It is important to note that the policy utilizes the original predicted scores $\hat{p}$ as input, rather than the calibrated scores, to maintain consistency with the subsequent RL training stage.

\subsubsection{On-Manifold Mixup Regularization}\label{sec:appendix_mixup}
The mixup regularization technique was initially introduced for supervised learning tasks \citep{zhang2017mixup}, where new input-output pairs are generated by taking convex combinations of pairs of training samples. \citet{wang2020improving} extended this approach to RL, where observations and their associated supervision signals from two transitions are combined convexly. In our case, the observation corresponds to the prompt embeddings. However, naively combining two prompt embeddings may produce vectors that lie outside the prompt manifold. To address this, we use the nearest neighbor from the replay buffer for each prompt $x$. Given the embedding $e$ for prompt $x$ and the embedding $e_n$ for its nearest neighbor, the interpolated prompt embedding is obtained as:
\begin{equation}
    \hat{e} = \lambda e + (1 - \lambda) e_n,
\end{equation}
where $\lambda \sim \text{Beta}(\xi, \xi)$, and $\xi$ is a hyperparameter, set to 0.2 as recommended in the original mixup paper. To train the routing policy on the interpolated prompt embeddings using PPO, we similarly interpolate the associated supervision signals:
\begin{equation}
\begin{aligned}
    \hat{\pi}_{old} &= \lambda \pi_{old} + (1 - \lambda) \pi_{old}^{(n)}\\
    \hat{\mathbf{A}} &= \lambda \mathbf{A} + (1 - \lambda) \mathbf{A}_n\\
    \hat{\mathbf{V}}_{targ} &= \lambda \mathbf{V}_{targ} + (1 - \lambda) \mathbf{V}_{targ}^{(n)}
\end{aligned}
\end{equation}
The interpolated routing action $\hat{a}$ is chosen as $a$ if $\lambda > 0.5$, otherwise $a_n$. Similarly, routing-relevant parameters, including $\mathbf{I}_k$ and $\boldsymbol{\omega}$ are chosen based on $\lambda$ as well.

\subsubsection{Reward Normalization}\label{sec:appendix_reward_normalization}
Our routing policy is designed to generalize across different sets of LLM candidates. However, the varying score and cost scales across these sets can pose challenges. For instance, routing decisions involving proprietary API models often involve higher costs compared to open-source models, where the cost is significantly lower. These discrepancies in scale can complicate the training of the routing policy, as the preference vector must be adjusted to suit each scenario. Moreover, the same preference vector might favor higher costs for one set of models while preferring lower costs for another, introducing inconsistency and instability during training. To address this, we propose normalizing both the scores and costs across all LLM sets. Given a set of LLMs $\{M_k\}_{k=1}^K$ with scores  $\{s_k\}_{k=1}^K$ and costs $\{c_k\}_{k=1}^K$, we normalize the scores and costs by
\begin{equation}
    \bar{s}_k = s_k / \max(\{s_k\}_{k=1}^K), \quad \bar{c}_k = c_k / \max(\{c_k\}_{k=1}^K).
\end{equation}
This normalization ensures that both scores and costs are scaled such that their maximum value is $1.0$. By standardizing the range of values, the policy can learn a consistent mapping from user preferences to routing decisions across various LLM sets. This approach prevents the policy from disproportionately favoring either high-cost or low-cost models based purely on their relative scales, promoting more balanced decisions that accurately reflect trade-offs between performance and cost.

In theory, the preference vector $\boldsymbol{\omega}$ can take any value in the range of $[0, \infty)$. However, for simplicity, we define it as $\boldsymbol{\omega} = [1, \omega]$, fixing the preference weight for scores at 1 and only varying the weight for cost. When $\omega = 0$, the model selection prioritizes high scores regardless of cost, while $\omega = \infty$ indicates a preference for the lowest-cost model. In practice, we found that sampling $\omega$ from the range $[0, 2]$ effectively captures the Pareto front.

\subsubsection{Stratified Sampling}\label{sec:appendix_stratified_sampling}
Generalizing the routing policy to a new model $\tilde{M}$ requires to obtain its identity vector $\tilde{I}$, which captures the model's unique strengths and weaknesses. However, evaluating the model on all available prompts is often prohibitively expensive, especially when new models are frequently introduced. In order to reduce the evaluation cost, we propose selecting a subset of informative prompts that effectively assess the model's capabilities. Specifically, given a set of prompts $X = \{x_n\}_{n=1}^N$ and the binarized evaluation scores $Y_k = \{\bar{y}_{kn}\}_{n=1}^N$ for each available LLM $M_k$, we assess the difficulty of each prompt based on the average prediction accuracy across all models $M_k$, i.e.,
\begin{equation*}
    \psi_n = \mathbb{E}_k \left[-\bar{y}_{kn} \log p_{kn} - (1- \bar{y}_{kn}) \log (1 - p_{kn}) \right].
\end{equation*}
We then apply stratified sampling using the difficulty $\psi_n$ as the strata. The stratified sampling ensures the selected prompts covers a range of difficulties, from easy to hard, providing a more balanced and informative assessment of the model's strengths and weaknesses. Once the subsets $\tilde{X}$ is selected, the model identity vector is computed as:
\begin{equation*}\textstyle
    \tilde{I} = \argmin_{I} \mathcal{L}_{irt} + \mathcal{L}_{KL} = \argmin_{I} \left[ \mathbb{E}_{\tilde{x}} \left[-\bar{y}\log p - (1 - \bar{y}) \log (1 -p) \right] + \KL \left(q(\tilde{I}) \| p(\tilde{I}) \right) \right],
\end{equation*}
where $p = \text{sigmoid}(f(\tilde{e}, \tilde{I}))$, and $\tilde{e}$ is the prompt embedding for prompts $\tilde{x} \in \tilde{X}$. 

The stratified sampling approach described above can also be extended to sample prompts from pairwise comparison datasets. Given a pairwise comparison dataset $\mathcal{V}$, where each example consists of a prompt $x_n$, a pair of models $M_{k1}$ and $M_{k2}$, and a winning label $z_n \in \{0, 1\}$. We first assess each model's capability using Elo score \citep{elo1967proposed}. The Elo scores are then used as strata to sample a set of models as the comparison baselines. For each baseline $M_k$, we uniformly select a set of prompts $X_k$ on which to run inference with the new model $\tilde{M}$ and compare its performance to the baseline $M_k$. After obtaining the baseline models and pairwise comparison labels, the model identity vector is computed as:
\begin{equation}\textstyle
    \tilde{I} = \argmin_{I} \mathcal{L}_{pair} + \mathcal{L}_{KL}.
\end{equation}

In our experiments, we opted to sample prompts from existing evaluation benchmarks for simplicity. We leave the exploration of sampling from pairwise comparison datasets as future work.

\section{Training Algorithm}\label{sec:algorithm}

\begin{algorithm*}[h]
\caption{Training Procedure of Preference Conditioned Dynamic Routing}
\label{alg:routing}
\begin{algorithmic}[1]
\REQUIRE Model identify vectors $I$, Pretraining steps $T1$, Training steps $T2$ 
\REQUIRE Comparison dataset $\mathcal{V}$ for pretraining, Evaluation leaderboard $\mathcal{D}$ for training
\REQUIRE Evaluation score predictor $f(x, \mathbf{I}_k)$, Calibration parameters $\alpha$ and $\beta$
\REQUIRE Routing policy $\pi_{\theta}$, Preference range $[\omega_{min}, \omega_{max}]$, RL training procedure $\mathbb{P}$
\STATE \colorbox{gray!30}{\makebox[0.98\linewidth][l]{\textit{\#\# Pretraining Stage}}}
\FOR{step in $[1, \ldots, T1]$}
\STATE sample a batch of pretraining data $(x, (k1, k2), (c1,c2)) \sim \mathcal{V}$
\STATE sample a batch of preference $\boldsymbol{\omega} = [1, \omega]$ and $\omega \sim U(\omega_{min},\omega_{max})$ \COMMENT{uniform for cost} 
\STATE compute score predictions $\hat{p}_k = \text{sigmoid}(f(x,\mathbf{I}_k))$ \COMMENT{auxiliary info to the policy}
\STATE calibrate the score predictions $\bar{p}_k = \text{sigmoid}(\alpha f(x,\mathbf{I}_k) + \beta)$ \COMMENT{only used to predict action}
\STATE normalize scores $\bar{p}_k = \bar{p}_k / \max(\{\bar{p}_k\}_{k \in \{k1,k2\}})$ and costs $\bar{c}_k = c_k / \max(\{c_k\}_{k \in \{k1,k2\}})$
\STATE obtain routing action $\hat{a} = \argmax_{k \in \{k1,k2\}} \boldsymbol{\omega}^T [\bar{p}_k, -\bar{c}_k]$ \COMMENT{maximize scalarized reward}
\STATE pretrain the policy by optimizing $- \log \pi(\hat{a} \mid x, \{(\mathbf{I}_k, \bar{c}_k, \hat{p}_k)\}_{k \in \{k1,k2\}}, \boldsymbol{\omega})$
\ENDFOR
\STATE \colorbox{gray!30}{\makebox[0.98\linewidth][l]{\textit{\#\# Training Stage}}}
\STATE Initialize a replay buffer $\mathbb{B}$ \COMMENT{with on-manifold mixup regularization}
\FOR{step in $[1, \ldots, T2]$} 
\STATE sample a batch of training data $(x, \{(M_k, c_k, s_k)\}_{k=1}^K) \sim \mathcal{D}$ \COMMENT{$K$ is different across batches}
\STATE sample a batch of preference $\boldsymbol{\omega} = [1, \omega]$ and $\omega \sim U(\omega_{min},\omega_{max})$ \COMMENT{uniform for cost}
\STATE normalize scores $\bar{s}_k = s_k / \max(\{s_k\}_{k=1}^K)$ and costs $\bar{c}_k = c_k / \max(\{c_k\}_{k=1}^K)$
\STATE compute score predictions $\hat{p}_k = \text{sigmoid}(f(x,\mathbf{I}_k))$ \COMMENT{auxiliary info to the policy}
\STATE run the current policy $a \sim \pi(x, \{(\mathbf{I}_k, \bar{c}_k, \hat{p}_k)\}_{k=1}^K, \boldsymbol{\omega})$ and obtain reward $[\bar{s}_a, -\bar{c}_a]$
\STATE update replay buffer $\mathbb{B} \leftarrow (x, a, \{(M_k, \bar{c}_k, \bar{s}_k)\}_{k=1}^K, \boldsymbol{\omega})$ 
\STATE RL training on data sampled from the replay buffer $\mathbb{P}(\pi_{\theta}, \mathbb{B})$ \COMMENT{with mixup interpolation}
\ENDFOR
\end{algorithmic}
\end{algorithm*}

\section{Additional Related Works}\label{sec:appendxi_related_works}

\textbf{Generalization in Reinforcement Learning}\quad
Generalizing RL policies to new tasks, often referred to as zero-shot RL, is a growing area of research focused on enabling policies to handle unseen tasks without retraining \citep{korkmaz2024survey}. Approaches typically fall into three categories: The first category focuses on maximizing worst-case performance across tasks, often using adversarial training \citep{moos2022robust,dong2023robust}. This approach is commonly used when no data is available to identify the current task. The second category aims to compute task representations from data, allowing agents to adapt their policies to the specific task at hand. This approach is commonly employed in multi-task RL and hidden-parameter MDPs \citep{konidaris2014hidden}, where task representations are inferred from exploration data within the task environment \citep{touati2021learning,agarwal2021contrastive,benjamins2022contextualize,ingebrand2024zero}. The third category leverages in-context learning by feeding data from the current task directly into a pretrained transformer as context \citep{melo2022transformers,brohan2022rt}. Although transformers have demonstrated effectiveness, their high memory consumption, training instability, and data inefficiency present challenges to their broader application. Our routing policy falls into the second category, where the task representation is explicitly provided as a set of LLMs and their associated costs. In a similar vein, \citet{jain2020generalization} explore RL generalization to new action spaces using a VAE to learn action representations, whereas we capture LLM capabilities via identity vectors.

In addition to task generalization, research has also explored generalizing RL policies to new observation distributions. Techniques include data augmentation \citep{cobbe2019quantifying,yarats2021image,laskin2020reinforcement}, specialized architectures \citep{lee2019network}, regularization methods \citep{farebrother2018generalization,wang2020improving}, invariant representation learning \citep{tachet2018learning,zhang2020learning,agarwal2021contrastive}, and adversarial observation perturbations \citep{zhang2021generalization,korkmaz2022deep}. Our approach explores a simple regularization technique that encourage smoothness across prompt distributions.

\section{Experiment}\label{sec:appendix_experiment}

\subsection{Model Cost}\label{sec:appendix_model_cost}
In Table~\ref{tab:model_cost}, we list the costs for each model. For proprietary APIs, the costs are based on their official API pricing, while for open-source models, we reference pricing from TogetherAI\footnote{\url{https://www.together.ai/pricing}}. All costs are normalized by estimating the expense of processing 1 million input tokens and generating 1 million output tokens.

\begin{table}[]
    \centering
    \caption{The estimated cost of invoking the models for processing 1M input tokens and generating 1M output tokens.}
    \label{tab:model_cost}
    \begin{tabular}{c|c}
    \toprule
        Model & Cost (\$) \\
    \midrule
         gpt-3.5-turbo-0125 & 2 \\
         gpt-3.5-turbo-0301 & 3.5 \\
         gpt-3.5-turbo-0613 & 3.5 \\
         gpt-3.5-turbo-1106 & 3 \\
         gpt-4-0125-preview & 40 \\
         gpt-4o-2024-05-13 & 20 \\
         gpt-4o-mini-2024-07-18 & 0.75 \\
         gpt-4 & 90\\
         gpt-4-1106-preview & 40 \\
         gpt-4-turbo-2024-04-09 & 40 \\
         gpt-4-turbo & 40 \\
    \midrule
         claude-3-opus & 90 \\
         claude-3.5-sonnet & 18 \\
         claude-3-sonnet & 18 \\
         claude-3-haiku & 1.5 \\
         claude-2.1 & 32 \\
         claude-2 & 32 \\
         claude-instant & 3.2 \\
         claude-1 & 32 \\
    \midrule
         gemini-pro-1.5 & 14 \\
         gemini-flash-1.5 & 0.375 \\
    \midrule
         llama3.1-405b & 9 \\
         llama3.1-70b & 1.584 \\
         llama3.1-8b & 0.324 \\
         llama3-70b & 1.584 \\
         llama3-8b & 0.324 \\
    \midrule
         mistral-large & 12 \\
         mistral-medium & 10.8 \\
         mistral-small & 8 \\
         mixtral-8x22b & 2.16 \\
         mixtral-8x7b & 1.08 \\
         mixtral-7b & 0.36 \\
    \midrule
         command-r-plus & 18 \\
         command-r & 2 \\
         command & 3 \\
         command-light & 0.9\\
    \midrule
         qwen-1.5-110b & 3.24 \\
         qwen-1.5-72b & 1.62 \\
    \midrule
         yi-large & 6 \\
    \bottomrule
    \end{tabular}
\end{table}

\subsection{Dataset Statistics}
Our framework consists of three training stages: First, we train the IRT model to obtain model identity vectors $I$ and the evaluation score prediction model $f$. Second, we perform supervised pretraining of the routing policy on diverse prompts. Third, we train the routing policy using a reinforcement learning procedure. Below, we summarize the datasets used in each training stage.

The datasets used in this work fall into two categories: First, pairwise comparison datasets, where annotations indicate which of two models provides a higher-quality response. Second, LLM evaluation datasets, which provide evaluation scores for various models on a set of prompts. Table~\ref{tab:datasets} summarizes the statistics of these two types of datasets. We apply basic preprocessing, such as removing multi-turn prompts and excluding ties from pairwise comparisons. For LLM evaluation benchmarks, we select a subset of popular LLMs. Please see Table~\ref{tab:models} for the full list of LLMs involved in this work.

The IRT model is trained using the pairwise comparison datasets and the training splits of the evaluation datasets. The pretraining stage also uses these pairwise comparison datasets. For the policy training stage, the routing policy is trained separately on each LLM evaluation dataset. We do not train the policy across different evaluation benchmarks, as they employ different scoring mechanisms, leading to variations in score scales.

\begin{table}[]
    \centering
    \small
    \caption{Two types of datasets used in the training process.}
    \label{tab:datasets}
    \begin{tabular}{c|c|cc}
    \toprule
        Category & Dataset & \# Prompts & \# Models \\
    \midrule
        \multirow{5}{*}{Pairwise Model Comparison} & berkeley-nest/Nectar\tablefootnote{\url{https://huggingface.co/datasets/berkeley-nest/Nectar}} & 182954 & 39 \\
                                  & lmsys/lmsys-arena-human-preference-55k\tablefootnote{\url{https://huggingface.co/datasets/lmsys/lmsys-arena-human-preference-55k}} & 39716 & 64 \\
                                  & lmsys/chatbot\_arena\_conversations\tablefootnote{\url{https://huggingface.co/datasets/lmsys/chatbot_arena_conversations}} & 18320 & 20 \\
                                  & lmsys/mt\_bench\_human\_judgments\tablefootnote{\url{https://huggingface.co/datasets/lmsys/mt_bench_human_judgments}} & 894 & 6 \\
                                  & routellm/gpt4\_judge\_battles\tablefootnote{\url{https://huggingface.co/datasets/routellm/gpt4_judge_battles}} & 84864 & 2 \\
    \midrule
        \multirow{5}{*}{Single Model Evaluation} & AlpacaEval 2.0\tablefootnote{\url{https://tatsu-lab.github.io/alpaca_eval/}} & 805 & 61 \\
                                & HELM-Lite\tablefootnote{\url{https://crfm.stanford.edu/helm/lite/latest/}} & 13021 & 61 \\
                                & HELM-MMLU\tablefootnote{\url{https://crfm.stanford.edu/helm/mmlu/latest/}} & 14042 & 45 \\
                                & OpenLLM Leaderboard\tablefootnote{\url{https://huggingface.co/spaces/open-llm-leaderboard-old/open_llm_leaderboard}} & 14617 & 41 \\
                                & OpenLLM Leaderboard v2\tablefootnote{\url{https://huggingface.co/spaces/open-llm-leaderboard/open_llm_leaderboard}} & 21606 & 39 \\
    \bottomrule
    \end{tabular}
\end{table}

\begin{table}[]
\centering
\scriptsize	
\caption{The models used in this work for training and evaluating the routing policy.}
\label{tab:models}
\begin{tabular}{|c|c|c|c|}
\hline
ai21\_j2-grande & ai21\_j2-jumbo & ai21\_jamba-instruct & alpaca-7b \\
alpaca-13b & chatglm-6b & chatglm2-6b & chatglm3-6b \\
claude-1 & claude-2.0 & claude-2.1 & claude-instant-1 \\
claude-instant-1.2 & claude-3-5-sonnet-20240620 & claude-3-opus-20240229 & claude-3-sonnet-20240229 \\
claude-3-haiku-20240307 & cohere\_command-r & cohere\_command & cohere\_command-r-plus \\
cohere\_command-light & cohere\_command-xlarge & codellama-7b-instruct & codellama-13b-instruct \\
codellama-34b-instruct & codellama-70b-instruct & deepseek-llm-67b-chat & dolly-v2-12b \\
dolphin-2.2.1-mistral-7b & dialogpt-large & falcon-180b-chat & falcon-40b-instruct \\
falcon-7b-instruct & fastchat-t5-3b & flat-t5-small & gemini-1.0-pro \\
gemini-1.5-pro & gemini-1.5-flash & gemini-pro-dev-api & gemma-2-9b-it \\
gemma-2-27b-it & gemma-2b-it & gemma-7b-it & recurrentgemma-2b-it \\
recurrentgemma-9b-it & google-text-unicorn & google-text-bison & gpt2 \\
gpt2-large & gpt2-medium & gpt2-xl & gpt-3.5-turbo-0125 \\
gpt-3.5-turbo-0314 & gpt-3.5-turbo-0613 & gpt-3.5-turbo-1106 & gpt-4-0125-preview \\
gpt-4 & gpt-4-0314 & gpt-4-0613 & gpt-4-1106-preview \\
gpt-4-turbo-2024-04-09 & gpt-4o-2024-05-13 & gpt-4o-mini-2024-07-18 & gpt4all-13b-snoozy \\
guanaco-13b & guanaco-33b & guanaco-65b & guanaco-7b \\
koala-13b & llama-13b & llama-65b & llama-2-13b-chat \\
llama-2-70b-chat & llama-2-7b-chat & llama2-70b-steerlm-chat & llama-3-70b-instruct \\
llama-3-8b-instruct & llama-3.1-405b-instruct-turbo & llama-3.1-70b-instruct-turbo & llama-3.1-8b-instruct-turbo \\
luminous-base & luminous-supreme & luminous-extended & mamba-gpt-7b-v2 \\
metamath-13b & metamath-70b & mistral-7b-instruct-v0.1 & mistral-7b-instruct-v0.2 \\
mistral-7b-instruct-v0.3 & mistral-large & mistral-medium & mistral-small \\
mixtral-8x7b-instruct-v0.1 & mixtral-8x22b-instruct-v0.1 & mpt-30b-chat & mpt-7b-chat \\
nous-hermes-2-mixtral-8x7b-dpo & oasst-pythia-12b & opt-1.3b & opt-2.7b \\
opt-350m & opt-6.7b & opt-iml-max-1.3b & opt-iml-max-30b \\
openchat-3.5 & openchat-3.5-0106 & openhermes-2.5-mistral-7b & phi-2 \\
phi-2-dpo & phi-2-sft & phi-3-medium & phi-3-small \\
phi-3-mini & palm-2 & pythia-12b & pplx-70b-online \\
pplx-7b-online & palmyra-x-v3 & palmyra-x-v2 & qwen-14b-chat \\
qwen1.5-0.5b-chat & qwen1.5-1.8b-chat & qwen1.5-4b-chat & qwen1.5-72b-chat \\
qwen1.5-14b-chat & qwen1.5-32b-chat & qwen1.5-7b-chat & qwen1.5-110b-chat \\
qwen1.5-moe-a2.7b-chat & qwen2-0.5b-instruct & qwen2-1.5b-instruct & qwen2-7b-instruct \\
qwen2-72b-instruct & rwkv-4-raven-1b5 & rwkv-4-raven-3b & rwkv-4-raven-7b \\
rwkv-4-raven-14b & solar-10.7b-instruct-v1.0 & stablelm-tuned-alpha-7b & starling-lm-7b-alpha \\
stripedhyena-nous-7b & text\_davinci\_001 & text\_davinci\_002 & text\_davinci\_003 \\
tulu-2-dpo-7b & tulu-2-dpo-13b & tulu-2-dpo-70b & ultralm-13b \\
ultralm-65b & vicuna-13b & vicuna-33b & vicuna-7b \\
wizardlm-7b & wizardlm-13b & wizardlm-70b & yi-6b-chat \\
yi-34b-chat & yi-large & yi1.5-6b-chat & yi1.5-9b-chat \\
yi1.5-34b-chat & zephyr-7b-alpha & zephyr-7b-beta & \\
\hline
\end{tabular}
\end{table}

\subsection{Training the IRT Model}
The IRT model for evaluation outcome prediction consist of four component: the prompt representation $e$, the model identity vector $I$, the evaluation score predictor $f(e, I)$, and the pairwise winner predictor $g(e, I)$. The prompt representation $e$ is obtained using a pretrained text embedding model, meaning it contains no learnable parameters. The model identity vector is initialized as random embeddings for each model listed in Table~\ref{tab:models}, with the embedding dimension set to $128$. The two neural networks, $f$ and $g$, share a common backbone, differing only in their final linear layer. This shared architecture encourages the model identity vector to capture both types of evaluation outcomes, enabling more accurate representation of each model's strengths and weaknesses.

The IRT model is trained using both the pairwise comparison datasets and the training splits of the evaluation datasets, with a combined loss function, $\mathcal{L}_{irt} + \mathcal{L}_{pair}$. The model is trained for 10 epochs with a batch size of $256$. We use the Adam optimizer with a learning rate of $0.001$. The learning rate is decayed by $0.95$ after each epoch. We did not conduct extensive hyperparameter tuning, and no signs of overfitting were observed during preliminary experiments. Additionally, training beyond 10 epochs did not lead to further improvements in validation performance and downstream routing performance.

\subsection{Supervised Pretraining of the Routing Policy}
The supervised pretraining stage for the routing policy optimizes the following negative log-likelihood $ - \log \pi(\hat{a} \mid x, \{(\mathbf{I}_k, \bar{c}_k, \hat{p}_k)\}_{k \in \{k1, k2\}}, \boldsymbol{\omega})$ using the pairwise comparison dataset $\mathcal{V}$. Given two models $M_{k1}$ and $M_{k2}$ compared on the prompt $x$, we first sample a preference vector $\boldsymbol{\omega} = [1, \omega]$, where $\omega$ is uniformly sampled from the predefined distribution $U(\omega_{min}, \omega_{max})$. The routing decision is then estimated as $\hat{a} = \argmax_{k \in \{k1, k2\}} \boldsymbol{\omega}^T [\bar{p}_k, -\bar{c}_k]$, where $\bar{c}_k$ represents the normalized cost $\bar{c}_k = c_k / \max(c_{k1}, c_{k2})$, and $\bar{p}_k$ represent the calibrated evaluation score $\bar{p}_k = \text{sigmoid}(\alpha f(x, \mathbf{I}_k) + \beta)$. The evaluation scores are further normalized by dividing by the maximum calibrated scores, ensuring consistency with the scale used in the RL training stage. Note that these calibrated and normalized scores are used only for routing action estimation during pretraining, the policy takes in the original score prediction $\hat{p}_k = \text{sigmoid}(f(x,\mathbf{I}_k))$ as auxiliary inputs, since the normalized scores are not available during the test phase.

The pretraining stage runs for $500$ steps with a batch size of $1024$, using the Adam optimizer with a learning rate of $0.001$. The calibration parameters, $\alpha$ and $\beta$, are learned by fitting a logistic regression model. Again, we did not conduct extensive hyperparameter tuning, further tuning may improve the performance.

\subsection{RL Training of the Routing Policy}
The RL training stage follows a modified PPO procedure tailored for the multi-objective optimization task. Specifically, from the evaluation leaderboard $\mathcal{D}$, we sample $K$ models, $\{M_k\}_{k=1}^K$, as routing candidates. The costs $c_k$ of these models are normalized by $\bar{c}_k = c_k / \max(\{c_k\}_{k=1}^K)$, and the their evaluation scores $s_k$ are normalized by $\bar{s}_k = s_k / \max(\{s_k\}_{k=1}^K)$. The user preference $\boldsymbol{\omega} = [1, \omega]$ and $\omega$ is sampled from the distribution $U(\omega_{min}, \omega_{max})$. The training process starts with generating the trajectories following the current policy $a \sim \pi(x, \{(\mathbf{I}_k, \bar{c}_k, \hat{p}_k)\}_{k=1}^K, \boldsymbol{\omega})$. The multi-objective reward for action $a$ is represented as a vector $[\bar{s}_a, -\bar{c}_a]$. We update the replay buffer with these sampled trajectories and use samples from the buffer to train the policy. For mixup regularization, we identify the nearest neighbor for each sampled prompt and perform a weighted linear combination of the prompt embedding and its neighbors, where the weights are drawn from $\text{Beta}(0.2, 0.2)$. Both the reward and the advantage are linearly combined using the same weights.

At each training step, we sample $256$ new prompts, along with their routing candidates and preference vectors, to obtain the routing trajectories and update the replay buffer. The training stage runs for $500$ steps with a batch size of $256$, using Adam optimizer with learning rate of $0.001$. 

\subsection{Evaluation Setup}
We evaluate the routing performance on 5 LLM evaluation benchmarks and various sets of routing candidates. Table~\ref{tab:eval_setup} presents the detailed evaluation settings.

\begin{table}[]
    \centering
    \small
    \caption{Evaluation settings.}
    \label{tab:eval_setup}
    \begin{tabular}{c|c|c}
    \toprule
        Benchmark & Setting & Models \\
    \midrule
        AlpacaEval 2.0 & GPT4/Mixtral-8x7B & gpt4\_1106\_preview \\
                       &                   & Mixtral-8x7B-Instruct-v0.1 \\
    \cmidrule{2-3}
                       & GPT Family & gpt-3.5-turbo-0301 \\
                       &            & gpt-3.5-turbo-0613 \\
                       &            & gpt-3.5-turbo-1106 \\
                       &            & gpt-4-0125-preview \\
                       &            & gpt-4o-2024-05-13 \\
                       &            & gpt4 \\
                       &            & gpt4\_0314 \\
                       &            & gpt4\_0613 \\
                       &            & gpt4\_1106\_preview \\
    \cmidrule{2-3}
                       & Claude Family & claude \\
                       &               & claude-2 \\
                       &               & claude-2.1 \\
                       &               & claude-3-5-sonnet-20240620 \\
                       &               & claude-3-opus-20240229 \\
                       &               & claude-3-sonnet-20240229 \\
                       &               & claude-instant-1.2 \\
    \midrule
        HELM-MMLU & GPT4/Mixtral-8x7B & gpt4\_1106\_preview \\
                  &                   & mixtral-8x7b-32kseqlen \\
    \cmidrule{2-3}
                  & Mistral Family & mistral-7b-instruct-v0.3 \\
                  &                & mixtral-8x22b \\
                  &                & mixtral-8x7b-32kseqlen \\
    \cmidrule{2-3}
                  & GPT Family & gpt-3.5-turbo-0613 \\
                  &            & gpt-4-0613 \\
                  &            & gpt-4-1106-preview \\
                  &            & gpt-4o-2024-05-13 \\
    \midrule
        HELM-Lite & GPT4/Mixtral-8x7B & gpt4\_1106\_preview \\
                  &                   & mixtral-8x7b-32kseqlen \\
    \cmidrule{2-3}
                  & Mistral Family & mistral-7b-instruct-v0.3 \\
                  &                & mixtral-8x7b-32kseqlen \\
                  &                & mixtral-8x22b \\
    \cmidrule{2-3}
                  & GPT Family & gpt-4o-2024-05-13 \\
                  &            & gpt-4o-mini-2024-07-18 \\
                  &            & gpt-3.5-turbo-0613 \\
                  &            & gpt-4-0613 \\
                  &            & gpt-4-1106-preview \\
    \midrule
        OpenLLM   & Yi1.5 Family & Yi-1.5-34B-Chat \\
                  &              & Yi-1.5-6B-Chat \\
                  &              & Yi-1.5-9B-Chat \\
    \cmidrule{2-3}
                  & Mistral Family & Mistral-7B-Instruct-v0.2 \\
                  &                & Mixtral-8x22B-Instruct-v0.1 \\
                  &                & Mixtral-8x7B-Instruct-v0.1 \\
    \cmidrule{2-3}
                  & LLaMA3 Family & Llama-3-70B-Instruct \\
                  &               & Llama-3-8B-Instruct \\
    \midrule
        OpenLLMv2 & Yi1.5 Family & Yi-1.5-34B-Chat \\
                  &              & Yi-1.5-6B-Chat \\
                  &              & Yi-1.5-9B-Chat \\
    \cmidrule{2-3}
                  & Qwen2 Family & Qwen2-0.5B-Instruct \\
                  &              & Qwen2-1.5B-Instruct \\
                  &              & Qwen2-72B-Instruct \\
                  &              & Qwen2-7B-Instruct \\
    \cmidrule{2-3}
                  & LLaMA3 Family & Llama-3-70B-Instruct \\
                  &               & Llama-3-8B-Instruct \\
    \bottomrule
    \end{tabular}
\end{table}

\subsection{Baselines}
In this section, we describe the implementation details of the baseline methods.

\subsubsection{RouteLLM}
RouteLLM \citep{ong2024routellm} develops a model that predicts the winning label between a pair of LLMs and selects the model based on a threshold applied to the predicted probability. To account for varying user preferences, we evaluate RouteLLM using a range of different thresholds.

\subsubsection{Predictor}
The predicted evaluation scores $\hat{p}_k = \text{sigmoid}(f(x, \mathbf{I}_k))$ can be used directly to compute the scalarized reward for an LLM $M_k$ as $r_{\boldsymbol{\omega}}(x,k) = \boldsymbol{\omega}^T [\hat{p}_k, -c_k]$. The routing decision is then made by selecting $\hat{a} = \argmax_k r_{\boldsymbol{\omega}}(x,k)$.

\subsubsection{Random}
The random routing policy selects models based on predefined probabilities for each model. Different user preferences are reflected by adjusting these probabilities. However, when there are more than two LLM candidates, specifying the probabilities becomes non-trivial, so we omit the random baseline in these scenarios.

\subsubsection{Oracle}
The oracle routing policy selects the model based on the actual evaluation scores, making the routing decision as $\hat{a} = \argmax_k r_{\boldsymbol{\omega}}(x, k) = \argmax_k \boldsymbol{\omega}^T [s(x, k), -c_k]$, where $s(x, k)$ represents the true performance score for model $M_k$ on prompt $x$.

\subsubsection{PPO}
For each LLM candidate set and each user preference, we train a separate PPO routing policy to maximize the scalarized reward $r_{\boldsymbol{\omega}}(x, k) = \boldsymbol{\omega}^T [s(x, k), -c_k]$.

\subsection{Additional Evaluation Results}
\begin{figure}
    \centering
    \includegraphics[width=0.5\linewidth]{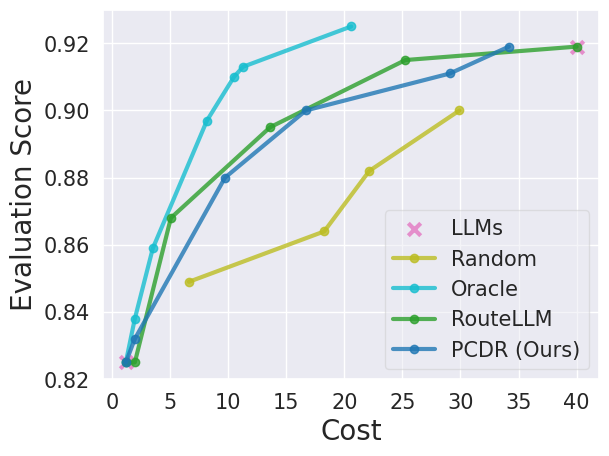}
    \caption{Performance-cost trade-off on MTBench dataset.}
    \label{fig:mt_bench}
\end{figure}

We also evaluate our preference-conditioned dynamic routing (PCDR) approach on MT-Bench, a widely-used benchmark for assessing LLM performance. Figure~\ref{fig:mt_bench} shows the performance-cost trade-off curves for different routing methods. While the Oracle policy achieves the best performance-cost trade-off as expected, our PCDR approach performs competitively with RouteLLM, particularly in the mid-to-high cost regime (\$20-40). Both methods significantly outperform random routing. Again, the gap between all routing policies and the Oracle baseline suggests potential room for improvement in routing decisions.

\begin{table}[]
    \centering
    \small
    \caption{Evaluation setting fro new routing candidates.}
    \label{tab:eval_setting_for_new_model}
    \begin{tabular}{c|c|c}
    \toprule
         Benchmark & Setting & Models \\
    \midrule
         OpenLLMv2 & Cohere & aya-23-35B \\
                   &        & aya-23-8B \\
    \cmidrule{2-3}
                   & Qwen2.5 & Qwen2.5-0.5B-Instruct \\
                   &         & Qwen2.5-1.5B-Instruct \\
                   &         & Qwen2.5-7B-Instruct \\
                   &         & Qwen2.5-14B-Instruct \\
                   &         & Qwen2.5-32B-Instruct \\
                   &         & Qwen2.5-72B-Instruct \\
    \bottomrule
    \end{tabular}
\end{table}

\subsection{Cold Start for New Routing Candidates}\label{sec:appendix_new_model}
To simulate the scenario where new models are introduced into the routing system, we select several unseen models from the HuggingFace OpenLLM v2 benchmark. These models are not used for training either the IRT model or the routing policy. Table~\ref{tab:eval_setting_for_new_model} shows the detailed evaluation settings.


\end{document}